\documentclass[Afour,sageh,times]{sagej}
\RequirePackage{fixltx2e}

\usepackage{moreverb,url}
\usepackage[colorlinks,bookmarksopen,bookmarksnumbered,citecolor=red,urlcolor=red]{hyperref}
\usepackage{amsmath}
\usepackage{algorithmic}
\usepackage{algorithm}
\usepackage{todonotes}
\usepackage{enumerate}
\usepackage{graphicx, caption, subcaption}
\usepackage{array}
\usepackage{eqparbox}

\DeclareMathOperator{\atantwo}{atan2}
\DeclareMathOperator{\sign}{sign}

\newcommand\BibTeX{{\rmfamily B\kern-.05em \textsc{i\kern-.025em b}\kern-.08em
		T\kern-.1667em\lower.7ex\hbox{E}\kern-.125emX}}

\def\IntSet{{I\!\!N}}
\def\volumeyear{2019}

\begin{document}

\runninghead{Cherubini et al. -- Model-free vision-based shaping of deformable plastic materials}

\title{Model-free vision-based shaping of deformable plastic materials}

\author{Andrea Cherubini\affilnum{1} Valerio Ortenzi\affilnum{2} Akansel Cosgun\affilnum{3} Robert Lee\affilnum{2} Peter Corke\affilnum{2}}

\affiliation{\affilnum{1}LIRMM, Universit\'{e} de Montpellier, CNRS, Montpellier, France.\\
\affilnum{2}ARC Centre of Excellence for Robotic Vision, Queensland University of Technology, Brisbane QLD 4001, Australia.  http://www.roboticvision.org \\
\affilnum{3}ARC Centre of Excellence for Robotic Vision, Monash University, Melbourne VIC, Australia.  http://www.roboticvision.org}

\corrauth{Andrea Cherubini, LIRMM, 860 Rue Saint Priest, 34090 Montpellier, France.} \email{andrea.cherubini@lirmm.fr}

\begin{abstract}
We address the problem of shaping deformable plastic materials using non-prehensile actions. Shaping plastic objects is challenging, since they are difficult to model and to track visually. We study this problem, by using kinetic sand, a plastic toy material which mimics the physical properties of wet sand. Inspired by a pilot study where humans shape kinetic sand, we define two types of actions: \textit{pushing} the material from the sides and \textit{tapping} from above. The chosen actions are executed with a robotic arm using image-based visual servoing. From the current and desired view of the material, we define states based on visual features such as the outer contour shape and the pixel luminosity values. These are mapped to actions, which are repeated iteratively to reduce the image error until convergence is reached. For pushing, we propose three methods for mapping the visual state to an action. These include heuristic methods and a neural network, trained from human actions. We show that it is possible to obtain simple shapes with the kinetic sand, without explicitly modeling the material. Our approach is limited in the types of shapes it can achieve. A richer set of action types and multi-step reasoning is needed to achieve more sophisticated shapes.
\end{abstract}


\maketitle

\section{Introduction}
\label{sec:introduction}

Many tasks such as cooking, folding clothes and gardening require the manipulation of soft objects and deformable materials. The same applies to industrial tasks such as inserting cables, packaging food and excavating soil and to medical ones such as injection, physiotherapy and surgery. Some of these tasks involve the manipulation of granular materials such as dough, sand, soil and salt. The capability of manipulating such materials would enable robots to perform a plethora of new applications that could increasingly help and substantially assist humans with their chores.

Despite the potential impact of successful manipulation of deformable materials, decades of robotics research have focused primarily on rigid objects. The reason behind this is the task difficulty. Adding to the inherent complexity of physical interaction with deformable materials are additional challenges, such as modeling the deformations. A direct consequence is that visual tracking of deformable materials is usually very demanding. 
Visual features of rigid objects can be consistently detected and tracked by exploiting a prior 3D model of the object. In contrast, features on deformable materials change over time and that can mislead both model-based or feature-based visual trackers. Feedback from force and tactile sensors could be beneficial, although this also requires deformation and contact models that map force/tactile signals to the corresponding displacements of the object surface. This mapping is commonly complicated to obtain.

With reference to the taxonomy given in the recent survey by~\cite{BoCoMe:18}, our focus is on \textit{solid} or \textit{volumetric} objects, \textit{i.e.}, objects with the three dimensions having comparable length. Sponges, plush toys, and food products fall in this category. In particular, a deformation occurs when an external force applied to an object changes its shape and appearance. Depending on the response of the object once the external force is removed, the deformation can be plastic, elastic, or elasto-plastic. Precisely, a \textit{plastic} deformation entails a permanent deformation, that is, an object maintains the shape caused by the applied force even when that force is removed.
 
 \begin{figure}[t!]
 	\centering 
 	\centering\includegraphics[width=\columnwidth]{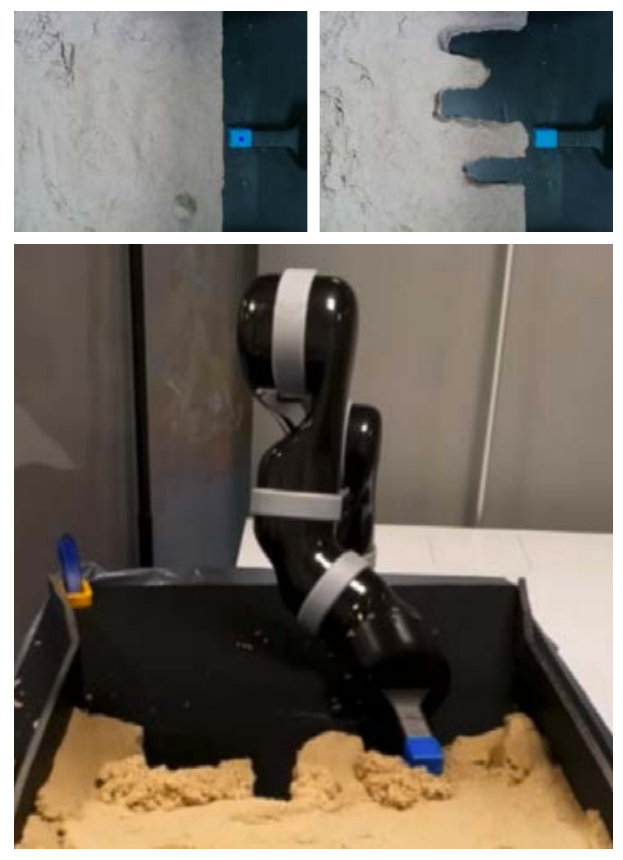}
 	\caption{Illustration of the \textit{shape servoing} problem that we address in this paper. Our goal is to find a sequence of non-prehensile manipulation actions a robot has to perform with a tool (blue in the figure) to mold a plastic material from an initial shape (top left) to a desired shape (top right). The bottom image shows the robot in action while it molds the material.}
 	\label{Fig:introFigure}
 \end{figure}

Humans often use non-prehensile actions, such as pushing or tapping, to move objects or modify the environment (see the work of~\cite{bullock2011classifying}). Such actions become even more common when the object of interest is plastic. For instance, the manipulation of clay or dough is typically non-prehensile. These observations motivated us to target non-prehensile actions.
 
In this work, we address the problem of \textit{shape servoing} plastic materials (see the works of~\cite{smolen2009deformation} and of~\cite{Na14}). Our goal is to \textit{find a sequence of non-prehensile actions a robot has to perform using a tool to bring plastic materials from an initial to a desired shape}. This problem is illustrated in Fig.~\ref{Fig:introFigure}.

To study plastic deformation, we use kinetic sand\footnote{\href{https://en.m.wikipedia.org/wiki/Kinetic_Sand}{https://en.m.wikipedia.org/wiki/Kinetic\_Sand}}, a toy material which mimics the physical properties of wet sand. It is made out of 98\% regular sand and 2\% polydimethylsiloxane (a viscoelastic silicone) and it can be molded into any desired shape. Kinetic sand does not stick to any materials other than itself and does not dry out. 

The paper is organized as follows. After reviewing the related literature (Sect.~\ref{Sect:relWork}) and presenting our contribution (Sect.~\ref{Sect:contributions}), we provide the results of a pilot study performed to understand how humans manipulate plastic materials (Sect.~\ref{sect:userstudy}). Then, in Sect.~\ref{Sect:problem} we define the shape servoing problem. Section~\ref{sect:method} describes our method, which is validated in the experiments of Sections~\ref{sect:setup} and~\ref{sect:experiments}. We conclude and propose directions for future work in Sect.~\ref{sect:conclusion}. 

\section{Related Work}
\label{Sect:relWork}

In this section, we review the literature on the manipulation of deformable objects, which is not as common as that of rigid objects. We also briefly review the literature on two tools needed in our work: visual servoing and action selection.

As observed in the survey on manipulating deformable objects by~\cite{BoCoMe:18}, much of the body of work focuses on 1D and 2D objects. Here, we mainly review works on solid 3D objects, as these are the ones addressed in our work. We first review shape estimation and tracking of deformable objects in Sect.~\ref{sec:rel1}, then shape control with and without a deformation model respectively in Sections~\ref{sec:rel2} and~\ref{sec:rel3}. Then, since in this paper we draw inspiration from Visual Servoing, we briefly recall the main works in this field (Sect.~\ref{sec:visServo}). Finally, we review methods on action selection (Sect.~\ref{sec:actionSel}).

\subsection{Shape Estimation/Tracking of deformable objects}
\label{sec:rel1}

The works that estimate and track deformable objects shape can be classified in three categories: those that require a model (generally mechanical), the data-driven ones (model-free) and the ones that combine the two.


Mechanical model-based trackers use a priori knowledge of the object's physical model. 
In one of the earliest papers on vision-based deformable shape estimation by~\cite{sarata2004trajectory}, stereo vision is used to obtain the 3D volumetric model of a pile for robotic scooping. \cite{petit2017tracking} use a Finite Element Method (FEM) for mesh-fitting. The goal is to track the shape of a pizza dough in real time. A volumetric FEM model is also used by~\cite{FrScSt:14}. An object is probed and the error between observed and simulated deformation is minimized to estimate its Young's modulus and Poisson ratio. \cite{guler2015estimating} estimate the deformations of elastic materials using optical flow and mesh-less shape matching.

Data-driven shape estimation methods rely on data gathered by sensors. \cite{khalil2010visual} track the surface of an object to be grasped by a robot using tactile and visual data. \cite{cretu2010deformable} address a similar application, by extracting the foreground before detecting the object contour. \cite{staffa2015segmentation} train a neural network for visual segmentation of non-rigid deformable objects (a pizza dough in their work).

Some approaches combine mechanical (model-based) and data-driven (model-free) estimation methods. \cite{caccamo2016active} use tactile and RGB-D data to create a mapping of how the surface is deformed, using Gaussian Processes. \cite{arriola2017multi} propose a generative model that uses force and vision to predict not only the object deformation, but also the parameters of a spring-mass model that represents it.

\subsection{Model-based Deformation Control}
\label{sec:rel2}

Some approaches take advantage of the deformable object's physical model, whenever it is available. \cite{howard2000intelligent} train a neural network on a physics-based model to extract the minimum force required for 3D object manipulation. \cite{gopalakrishnan2005d} focus on 2D objects and use a mesh model with linear elastic polygons. Their approach, called ``deform closure'' is an adaptation of ``form closure'', a well-known method for rigid object grasping. \cite{DaSa:2011} use a mass-spring-damper model to simulate
a planar object so that its shape, described by a curve, can be changed into another
desired shape. Their method, however, requires a significant number of actuation points (over 100), and is only validated in simulations. \cite{higashimori2010active} use a four-element model and present a two-step approach where the elastic parameters are estimated by force sensing, and then the force required to reach the desired shape is calculated based on the plastic response. \cite{cretu2012soft} monitor shape deformation by tracking lines that form a grid on the object. A feed-forward neural network is used to segment and monitor the deformation. \cite{arriola2017multi} predict the object behavior by first classifying the material, then using force and computer vision to estimate its plastic and elastic deformations. In~\cite{CoDuFi:18}, a method for dexterous in-hand manipulation of 3D soft objects for real-time deformation control is presented, relying on Finite Element Modeling. However, the authors assume the object to be purely elastic.

\subsection{Model-free Deformation Control}
\label{sec:rel3}

Some approaches explored deformable object manipulation without explicitly modelling the deformations. In their pioneering work,~\cite{WaHiKaKa01} designed a PID controller that can manipulate 2D objects in the absence of a prior model. \cite{smolen2009deformation} use a mesh-less model of the object where a set of points are controlled on the surface. A Jacobian transform is derived and used to control the robot motion. \cite{berenson2013manipulation} uses the concept of diminishing rigidity to compute an approximate Jacobian of the deformable object. In his work, human and robot simultaneously manipulate the object (a 2D cloth).

More recent model-free approaches rely on machine learning. \cite{gemici2014learning} learn haptic properties such as plasticity and tensile strength of food objects. Other researchers propose to predict the next state given the current state and a proposed action. For instance, \cite{elliott2018robotic} use some defined primitive tool action for rearranging dirt and \cite{schenck2017learning} present a Convolutional Neural Network for scooping and dumping granular materials. \cite{li2018learnParticle} propose to learn a particle-based simulator for complex control tasks. This enables the simulator to quickly adapt to new
environments or to unknown dynamics within a few observations. 

\subsection{Visual servoing}
\label{sec:visServo}

Visual servoing is a technique which uses visual features (e.g., points, lines, circles, etc.) extracted from a camera to control the motion of a robot. Visual servoing methods are commonly used in robot manipulation and can be classified as position-based or image-based. In position-based visual servoing, the feature is reprojected in the 3D space and the robot is controlled in Cartesian coordinates as done by~\cite{drummond1999visual}. Instead, the image-based approach regulates an error defined in the image space. This is done via the interaction matrix, which relates the dynamics of the camera (i.e., its velocity) to those of the visual feature to be controlled, as explained by~\cite{ChHu06a}. Image-based visual servoing is robust to calibration errors as shown by~\cite{HuHaCo96}. It can be used with various features, including image moments as done by~\cite{ChTa:05}, and mutual information by~\cite{Sh:48}, as done by~\cite{DaMa:2009}. The interested reader can refer to the works of~\cite{kragic2002survey} for a detailed survey, and to those of~\cite{ChHu06a,ChHu07} and~\cite{HuHaCo96} for tutorials on visual servoing.

Visual servoing can also be used for deformation control. For instance,~\cite{Na13, Na14} actively deform compliant objects using a novel visual servoing scheme that explicitly deals with elastic deformations, by estimating online the interaction matrix relating tool velocities and optical flow. Their controller is model-free, but focuses mainly on shape control, \textit{i.e.}, on manipulating the object to a desired configuration, without dealing with its global deformation over a long time window. The latest approach by \cite{navarro2018fourier} uses a Fourier series as visual feature for representing the object contour.~\cite{zhu:iros18} use a similar feature for dual arm shaping of flexible cables.

\subsection{Action selection methods}
\label{sec:actionSel}

When manipulation requires various types of actions (as is often the case with deformable objects) an automatic action selection method is needed. Although we did not implement any of these methods (in our work, the operator manually selects the action at each iteration), it is worth reviewing them, as they could be integrated in our framework. Such methods are often based on machine learning or on motion planning. 

Learning-based methods are frequently used for choosing within a finite number of actions. An example is deep reinforcement learning (see the works of~\cite{dulac2015deep} and of~\cite{isele2018navigating}), which relies on the Markov Decision Process (MDP) formulation (refer to~\cite{puterman2014markov}).~\cite{laskey2017learning} learn a policy for choosing actions to manipulate cloth using imitation learning.

~\cite{KaMo:06} present a planning method for finding intermediate states and transitioning between these states to shape a flexible wire. Similarly,~\cite{zhu:ral18} plan the sequence of actions for shaping a cable using environmental contacts.

\begin{figure*}[h!]
	\centering 
	\centering\includegraphics[width=\textwidth]{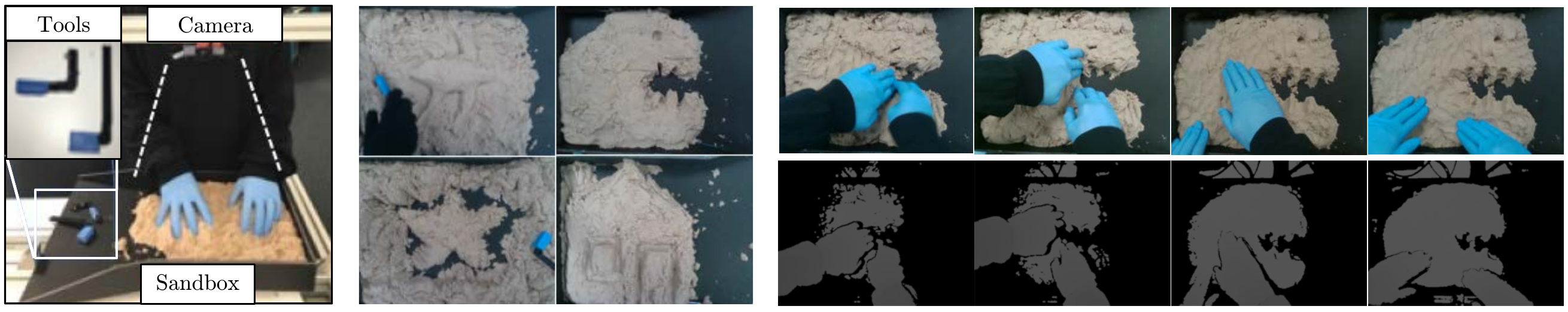}
	\caption{Human plastic material shaping study. Left: setup for the user study: an RGB-D camera (RealSense) is fixed above the sandbox, where participants mold the kinetic sand into a desired shape; first with one, then with two hands and third, using one of two provided tools. Center: four different shapes formed by the participants of our user study. Right: eight images (top: RGB, bottom: depth) acquired while a user is shaping the kinetic sand using both hands.}
	\label{Fig:setup}
\end{figure*}

\section{Contributions}
\label{Sect:contributions}


Our approach is model-free, because we do not model the deformations of the plastic material, like some of the other work reviewed in Sect.~\ref{Sect:relWork}. We instead focus on altering the visible characteristics of the material as in  \cite{navarro2018fourier}. Differently from the cited work, we make use of the data collected from humans for both the heuristic and learning-based algorithms explored in this paper. Most works in the field focus on a single deformation action, whereas our algorithmic choices are driven by the sequential nature of the task -- typically a sequence of non-prehensile actions are required to achieve a given desired shape. We also have a particular emphasis on the real-world implementation. 

In our previous work (see~\cite{ChCoOr:18}) we ran a user study on kinetic sand manipulation, presented the corresponding image dataset and designed an image processing algorithm for extracting features from the dataset. We also proposed an optimization-based algorithm for controlling the robot, along with a neural network architecture for mapping the state of the material to the pushing action to be applied. There were no robot experiments in that paper. 

The first three points (user study, dataset and image processing algorithm) constitute the foundations of the present work (see Sections~\ref{sect:userstudy} and~\ref{sect:gettingTriplets}). Nevertheless,~\cite{ChCoOr:18} did not address real experiments. As soon as we implemented them, the proposed tools/methods were confronted to their limitations. First, the user dataset was inappropriate for transferring manipulation capabilities from humans to robots, and we had to enrich it with extra data (see Sect.~\ref{Ref:userDis}, point (c)). Second, we had to completely redesign the problem statement and method (see Sect.~\ref{Sect:problem}). For instance, we had to address the pushing problem locally (see Sect.~\ref{sect.localShaping}), since the global method failed. Third, we modified the neural network architecture, and, given its limitations, compared it with two other strategies (\textit{maximum} and \textit{average}) which were not present in the previous work. Finally, we realized a series of unprecedented robotic experiments on shape servoing of plastic materials. 

In summary, the contributions of this work are:
\begin{enumerate}
	
	
	
	\item We exploit the human demonstration data to compensate for the lack of a physical material model. The human data is utilized to design a local strategy for robotic kinetic sand manipulation and also to train a machine learning-based action model.
	
	
	\item We propose two classes of non-prehensile actions to reduce the complexity of the deformable object manipulation problem: \textit{pushing} to modify the outer contours, and \textit{tapping} to adjust the kinetic sand height. 
	
	\item We report experiments where a robot manipulator successfully molds the kinetic sand into various desired shapes. In the experiments, we compare various approaches for realizing the pushing and tapping actions.
\end{enumerate}
In a nutshell, while Sections~\ref{sect:userstudy} and~\ref{sect:gettingTriplets} had been presented in our previous work (see~\cite{ChCoOr:18}), the rest of the present paper reports unpublished, original research.


\section{User study}
\label{sect:userstudy}

\subsection{Materials and methods}

To understand how humans manipulate plastic materials, we ran a pilot user study with 9 volunteers (age range: 20-40; 6 male, 3 female). Each participant was asked to shape kinetic sand in a sandbox, while being recorded with a fixed RGB-D camera (Intel RealSense SR300, resolution $640 \times 480$). We opted for such a low cost and easily available sensor to make the setup reproducible and inexpensive. The RealSense was pointing at the sandbox from above, with its optical axis perpendicular to it, as shown in Fig.~\ref{Fig:setup} (left). 

Each participant was requested to produce a shape of their own choice, repeating the task three times: a) using both hands, b) using only one hand and finally c) using one of the two provided tools (Fig.~\ref{Fig:setup}, left). 
Four examples of shapes formed by the participants are shown in Fig.~\ref{Fig:setup} (center). Figure~\ref{Fig:setup} (right) shows some of the images acquired while a user was shaping the kinetic sand with both hands.

All participants gave their consent to be recorded. We obtained ethics clearance to make the dataset (almost 214000 RGB and depth images) publicly available\footnote{\href{https://cloudstor.aarnet.edu.au/plus/s/Vii90T72WFM8Qwp}{https://cloudstor.aarnet.edu.au/plus/s/Vii90T72WFM8Qwp} (password: sandman)}. Afterwards, the participants filled out a questionnaire, to help us infer their ``sculpting strategy''. 
None of the participants had previous direct experience with sculpting, although 5 declared to have previously manipulated deformable objects such as dough or clay. 

\subsection{Results}
\label{Sect:results}

Six participants stated that they had performed a clear sequence of different actions while using their hands, while the number raised to eight when using the tool. In particular, the participants clearly identified \textit{pushing} as an action type. In our opinion, this is due to the fact that pushing has a very clear outcome (the kinetic sand moving and consequently changing its outer contours). Two other actions commonly identified were \textit{tapping} and \textit{incising}. 

The questionnaire included the following four questions: 
\begin{itemize}
	\item 
	``How much did you rely on vision while hand modeling? grade between 1 (very little) and 5 (very much)''
	\item 
	``How much did you rely on haptic feedback while hand modeling? grade between 1 (very little) and 5 (very much)''
	\item 
	``How much did you rely on vision while sculpting with the tool? grade between 1 (very little) and 5 (very much)''
	\item 
	``How much did you rely on haptic feedback while sculpting with the tool? grade between 1 (very little) and 5 (very much)''
\end{itemize}
All of the participants, whether using their hands or the tool, reported having relied on vision either ``very much'' or ``much''. Haptic feedback was only partially important during the trials (7 participants reported having relied ``little'' on haptic feedback). This result led us to believe that -- although the kinetic sand offers some force resistance while sculpting -- haptics is not as valuable a feedback as vision. In our opinion, it is more natural to rely on visual feedback as a measure of how distant the current shape is from the imagined/desired shape. Haptic feedback is valuable \textit{locally} in understanding the force required to overcome the kinetic sand resistance, but offers little \textit{global} information on which action to perform next, to obtain the desired shape. In summary, visual and haptic feedback are complementary. To further investigate the importance of haptics, one could blindfold the participants (so they cannot rely on vision) or have them teleoperate a molding robot without haptic feedback (so they can only rely on vision). These experiments could be the object of future work.

Since we focus on non-prehensile kinetic sand manipulation with a robotic tool, within the dataset we manually labeled only the images where participants used either tool. Within these, we only labeled the RGB images, since the RealSense depth image quality is insufficient for feature extraction (see Fig.~\ref{Fig:setup}, bottom right). Among the 53,100 RGB images of tool shaping, 13.3\% were labeled as ``pushing'', 9.3\% as ``tapping'' and 8.7\% as ``incising''. The other images were left unlabeled (68.7\%), as they present: retreating actions, transitions between actions, occlusions (\textit{i.e.}, the tool was not visible), simultaneous ``push-and-tap'' actions (the participant modifies the contour, while also yielding a smoothing effect on the surface) and unclear actions. 


\subsection{Discussion}
\label{Ref:userDis}
The results of the user study motivated us to adopt the following choices for transferring shape servoing capabilities from humans to robots:

\begin{enumerate}[a)]
		
	\item 
The importance of visual feedback prompts us to use a camera as the only source of feedback. This choice also enables us to use \textit{Image-Based Visual Servoing} (IBVS, see the works of~\cite{HuHaCo96} and~\cite{ChHu06a,ChHu07}) to control the robot directly in the image space, even with possibly coarse camera calibration.	

	\item
The results highlight that participants commonly rely on -- and can clearly recognize -- a \textit{finite set of actions} to model the kinetic sand. This motivates us to implement the two actions that are most heavily identified and chosen in the user study: \textit{pushing} and \textit{tapping}. Pushing shapes the outer contours of the kinetic sand, whereas tapping regularizes the surface by leveling it. Both are tool motions along a vector going towards the kinetic sand from the free space. For pushing, the vector is parallel to the sandbox plane, whereas for tapping it is orthogonal to the sandbox plane and pointing down. We discarded \textit{incising}, since it is the least popular among users. 
In our opinion, incising can be achieved by tracking a desired  trajectory (the incision) with the tool.
	
	\item
The user dataset is inappropriate for transferring the mentioned pushing and tapping capabilities from humans to robots. Only 12,000 labeled images (22.6\% of those in the tool dataset) are available. This quantity is insufficient for learning such complex actions. Hence, we decided to enrich the dataset with ad-hoc trials where participants performed only one of the two actions at a time. In each trial, the participant performed a sequence of ``clean'' (without occlusions and with clear movements) pushing or tapping actions. This simplified the image processing steps needed to produce the training dataset. In the following, we denote the complete collection of pushing and tapping images as the \textit{PT-Dataset}. The dataset contains 24,000 images (12,000 from the users and 12,000 from these ad-hoc trials), of which 14,900 labeled as ``pushing'' and 9,100 as ``tapping''.

\end{enumerate}

\section{Problem Statement}
\label{Sect:problem}

In this section, we define the shape servoing task and the work assumptions. The variables defined in this section are shown in Fig.~\ref{Fig:actions_states}.

\subsection{The shape servoing task}
\label{sect:shapeservo}

We consider the same setup as in the user study (Sect.~\ref{sect:userstudy}): a camera points at the object (here, the kinetic sand) from above, with a vertical optical axis.
Given a desired image $\mathbf{I}^*$ (e.g., one of those in Fig.~\ref{Fig:setup}, center), the \textit{shape servoing task} consists of shaping the kinetic sand until it appears as in $\mathbf{I}^*$.

We model the robot and object shape as a discrete-time system. 
The robot tool center point (TCP) can be moved in the workspace to modify the object shape.
At each iteration $k \in \left[  1, K \right] \subset \IntSet$, the robot observes the shape in image $\mathbf{I}_k$ (of width $w$ and height $h$), and then modifies it by moving its TCP to perform some \textit{action} $\mathbf{a}_k$. 

\textit{\underline{Definition: shape servoing task.}} We define $e_k \left(\mathbf{I}_k, \mathbf{I}^*\right) \geq 0$ as a scalar function that measures the \textit{image error} between $\mathbf{I}_k$ and $\mathbf{I}^*$, so that $e_k = 0 \iff \mathbf{I}_k = \mathbf{I}^*$. Then, the task of shaping an object to $\mathbf{I}^*$ with an accuracy $\bar{e} \geq 0$, consists in applying a finite sequence of actions $\mathbf{a}_1, \dots, \mathbf{a}_K$ such that after $K$ iterations: $e_K  \leq \bar{e}$. We tolerate such an upper bound on the accuracy, since even a human would be incapable of perfectly reproducing ($e_K  = 0$) a given image with the kinetic sand. Nevertheless, the sequence of actions should make $e_K$ decrease. More formally, it must be possible to apply, at each iteration $k$, an action
\begin{equation}
\mathbf{a}_k = \mathbf{a}\left( \mathbf{I}_k, \mathbf{I}^* \right)
\label{eq:actionImg}
\end{equation}
that will reduce $e_k$:
\begin{equation}
e_{k+1} < e_k.
\label{eq:errorDecrease}
\end{equation}

Inspired by the results of the user study, we consider a set of two actions, defined below.

\begin{figure}[t!]
 	\centering 
 	\centering\includegraphics[width=\columnwidth]{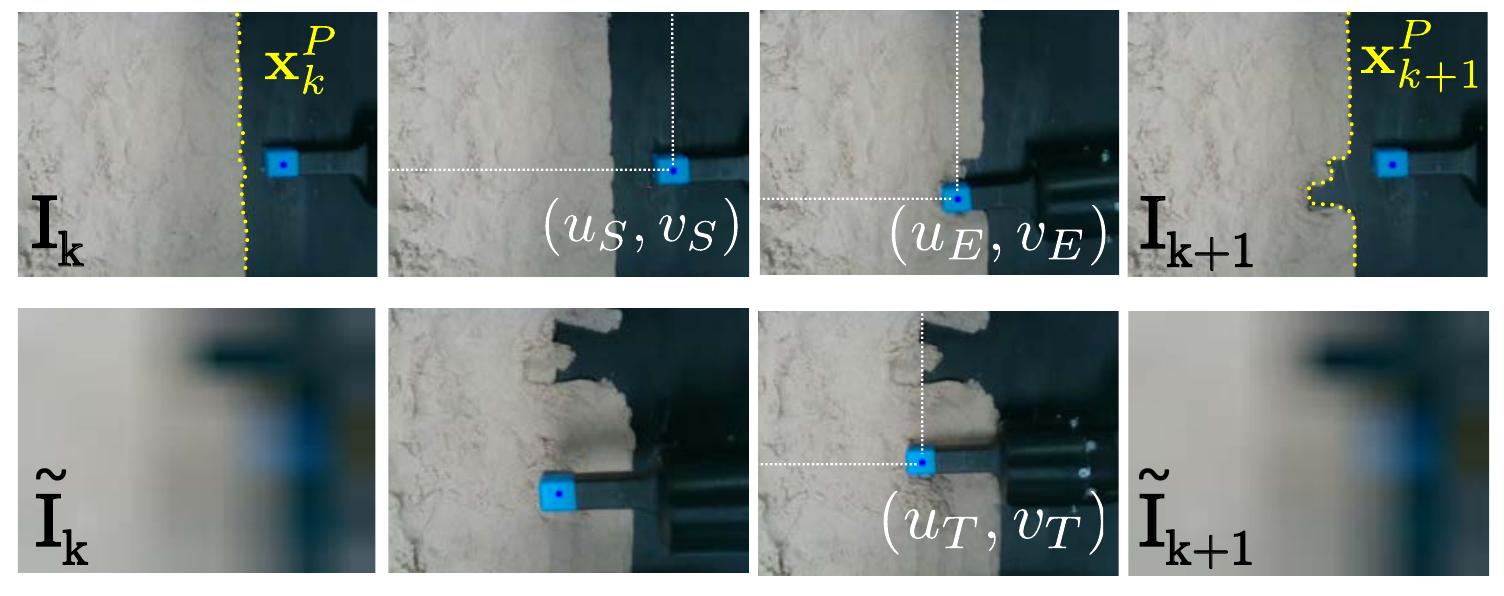}
 	\caption{The two non-prehensile actions that we use and their effect on the kinetic sand. Top: the pushing action ${\mathbf{p}} = \left[ u_S \; v_S \; u_E \; v_E \right]^\top$ modifies the outer contours (yellow) from ${\mathbf{x}}^P_k$ (on image $\mathbf{I}_k$) to ${\mathbf{x}}^P_{k+1}$ (on image $\mathbf{I}_{k+1}$). Bottom: the tapping action ${\mathbf{t}} = \left[u_T \; v_T \right]^\top$ adjusts the height and compactness of the kinetic sand; it also affects the resampled images $\tilde{\mathbf{I}}_k$ and $\tilde{\mathbf{I}}_{k+1}$ at the pixel $T$ where the tapping action is applied.}
 	\label{Fig:actions_states}
 \end{figure}

\textit{\underline{Definition: set of  actions.}} The robot can execute one of two actions, namely \textit{pushing} or \textit{tapping}: 
\begin{equation}
	\mathbf{a} = \left\{\mathbf{p},\mathbf{t}\right\},
	\label{eq:setActions}
\end{equation}
and each action is parameterized within a parameter set: $\mathbf{p} \in {\cal{A}}_P$, $\mathbf{t} \in {\cal{A}}_T$. 
We define the \textit{pushing} action ${\mathbf{p}}$ as a translation of the TCP between two points on the sandbox base. Indeed, if the contact between TCP and material is approximated by a point or by a sphere, the effect of pushing will be invariant to the tool orientation. Since the sandbox base is parallel to the image plane, ${\mathbf{p}}$ can be defined via the image coordinates $\left(u , v\right) \in \left[0, w\right] \times \left[0, h\right]$ of the start and end pixels (denoted $S$ and $E$, see top of Fig.~\ref{Fig:actions_states}) of the TCP motion:
\begin{equation}
	\begin{array}{l}
		{\mathbf{p}} = \left[ u_S \; v_S \; u_E \; v_E \right]^\top,\\
		{\cal{A}}_P\in \left[0, w\right]\times \left[0, h\right]\times\left[0, w\right]\times \left[0, h\right].
	\end{array}
	\label{eq:toolActionPushing}
\end{equation}
Similarly, we define the tapping action ${\mathbf{t}}$ as a vertical translation of the TCP, perpendicular to the sandbox base, between 2 points. Since the heights of these points are fixed, ${\mathbf{t}}$ can be parameterized only by the final position of the TCP in the sandbox plane, which corresponds to a pixel $T$ (see bottom of Fig.~\ref{Fig:actions_states}) in the image:
\begin{equation}
	\begin{array}{l}
		{\mathbf{t}} = \left[u_T \; v_T \right]^\top,\\
		{\cal{A}}_T\in \left[0, w\right] \times \left[0, h\right].
	\end{array}
	\label{eq:toolActionTapping}
\end{equation}
Other setups -- with the camera optical axis not perpendicular to the sandbox -- would require different representations.

\subsection{Assumptions}
\label{sec:assumptions}

In accordance with the user study (Sect.~\ref{Ref:userDis}), we make the following hypotheses. 

\textit{\underline{Hypothesis 1: specialised actions.}} 
Each action type $\mathbf{a} = \left\{\mathbf{p},\mathbf{t}\right\}$ regulates only some \textit{features} of image $\mathbf{I}_k$. This hypothesis follows from the user study, where the effects of the pushing and tapping actions were separate and distinct. Users push the object to shape its outer contour. In our setup, this feature is observable, since its perspective projection is also a contour, visible in the image and outlined in yellow in the top of Fig.~\ref{Fig:actions_states}. Then, the \textit{push-controlled feature} is the list of image coordinates of the $N$ pixels sampled along the shape's outer contour:
\begin{equation}
\mathbf{x}^P_k = \left[ 
u_1 \; v_1 \dots u_N \; v_N \right]^\top_k.
\label{eq:pushState}
\end{equation}
Similarly, when tapping, the users smooth and flatten the kinetic sand surface. This could be interpreted as regulating its height, a feature that is not measurable without accurate 3D sensing. To observe the effect of tapping on a monocular image, we draw inspiration from~\cite{ArRoSi:15}, where a robot flattens cloth by closing a feedback loop on the cloth wrinkles. Similarly, we measure the effect of tapping directly on image $\mathbf{I}_k$. We account for the tool resolution (\textit{i.e.}, the size of its contacting surface) by scaling $\mathbf{I}_k$ to a smaller $\tilde{\mathbf{I}}_k$, with pixels having the size of the tool's image when it contacts the kinetic sand: $w_{\mbox{TCP}} \times h_{\mbox{TCP}}$ pixels (of the original image). The \textit{tap-controlled feature} will correspond to this image $\tilde{\mathbf{I}}_k$ (shown in the bottom of Fig.~\ref{Fig:actions_states}) which can be obtained by linearly sampling ${\mathbf{I}}_k$ to a reduced size $w / w_{\mbox{TCP}} \times h / h_{\mbox{TCP}}$.

\textit{\underline{Hypothesis 2: feasible shape.}} Given the initial image $\mathbf{I}_1$, the desired image $\mathbf{I}^*$, and the set of actions defined in~(\ref{eq:setActions}), there exists at least one sequence of actions that solve the shape servoing problem defined above.
Practically, this hypothesis means that the desired shape can be realized using the set of actions defined in~(\ref{eq:setActions}), e.g., matter cannot be added nor pulled, etc.

Even if a solution exists, finding the best sequence of actions $\mathbf{a}^*_k$ that verifies~(\ref{eq:errorDecrease}) requires solving -- at each iteration $k$ -- two subproblems. First, one must find the best action within the set $\left\{\mathbf{p},\mathbf{t}\right\}$ and then, for the chosen action $\mathbf{a}$, one must find the best parameters within the action's parameter set -- either ${\cal{A}}_P$ or ${\cal{A}}_T$.

 \textit{\underline{Hypothesis 3: action type pre-selected at each iteration.}}  In this work, we only address the second subproblem: we assume that at each iteration either a cognition layer or a human operator has selected the best action type (\textit{push} or \textit{tap}), and we focus on finding the best parameters for \textit{that} selected action (\textit{where} to push or tap). 
 
 Despite these hypotheses, two non-trivial issues are still to be addressed: the choice of image error measure $e_k$ and the visual control strategy for defining $\mathbf{a}_k = \mathbf{a}\left( \mathbf{I}_k, \mathbf{I}^* \right)$. These choices will be the discussed in the next Section.

\section{Proposed method}\label{sect:method}
\subsection{Image error}

Since images $\mathbf{I}_k$ and $\mathbf{I}^*$ are spatially aligned into the same geometric base, we can design image error $e_k$ using a global similarity measure (see~\cite{Mitchell:2010}). These measures include mean square error, mean absolute error, cross-correlation, and mutual information (defined by~\cite{Sh:48}). Among the four, the latter is best in terms of robustness to light variations and occlusions.
The mutual information between images $\mathbf{I}_k$ and $\mathbf{I}^*$ is:
\begin{equation}
{\mbox{MI}}_k = {\mbox{MI}} \left(\mathbf{I}_k, \mathbf{I}^*\right) = \sum_{i_k,i^*} p\left(i_k, i^*\right) \log \left( \frac{p\left(i_k, i^*\right)}{p\left(i_k\right) p\left(i^*\right)} \right),
\label{eq:mutInfo}
\end{equation}
with:
\begin{itemize}
	\item 
	$i_k$ and $i^*$ the possible pixel values in images $\mathbf{I}_k$ and $\mathbf{I}^*$ (e.g., in the case of 8-bit grayscale images these are the luminances  and $\left(i_k, i^*\right) \in [0, 255] \times [0, 255]$),
	\item
	$p\left(i_k\right)$ and $p\left(i^*\right)$ the probabilities of value $i_k$ in $\mathbf{I}_k$ and $i^*$ in $\mathbf{I}^*$ (respectively),
	\item
	$p\left(i_k, i^*\right)$ the joint probability of $i_k$ and $i^*$ computed by normalization of the joint histogram of the images.
\end{itemize} 
${\mbox{MI}}_k$ is maximal and unitary when $\mathbf{I}_k \equiv \mathbf{I}^*$: ${\mbox{MI}} \left(\mathbf{I}^*, \mathbf{I}^*\right) = 1$. Then, to obtain $e_k = 0$ for $\mathbf{I}_k = \mathbf{I}^*$ and $e_k$ monotonically increasing with the image dissimilarity, we should set: 
\begin{equation}
e_k = {\mbox{MI}} \left(\mathbf{I}^*, \mathbf{I}^*\right) - {\mbox{MI}}_k = 1 - {\mbox{MI}}_k.
\label{eq:mutInfoError}
\end{equation}
In the following, we denote this expression of $e_k$ as the \textit{mutual information error}.

\subsection{Image-based control}
\label{sec:image_based_control}
Having chosen $e_k$, the second issue is the definition of a robot control input $\mathbf{a} \left( \mathbf{I}_k, \mathbf{I}^* \right)$ that ensures~(\ref{eq:errorDecrease}) when $e_k = {\mbox{MI}}_k$.  
\cite{DaMa:2009} address such problem (Mutual Information-based visual servoing) by applying Levenberg-Marquardt optimization to maximize ${\mbox{MI}}$. Yet, they can derive analytic expressions of both the Gradient and Hessian of ${\mbox{MI}}$ with respect to $\mathbf{a}$, since in their work $\mathbf{a}$ is the pose of the camera looking at a photography (the image).
Conversely, in our application we do not have an accurate model of the dynamics of the scene, which evolves, is not rigid and cannot be assumed Lambertian, as in the work of~\cite{DaMa:2009}. Hence, we cannot derive an analytic relationship between the applied action $\mathbf{a}$ and the corresponding dynamics of ${\mbox{MI}}$ and therefore find some $\mathbf{a}^*$ that guarantees~(\ref{eq:errorDecrease}). Thus, while ${\mbox{MI}}$ is an objective metric for assessing the controller convergence, it is not straightforward to use in the design of $\mathbf{a}$. 

Instead, based on \underline{\textit{Hypothesis 1}}, we design $\mathbf{a}$ based on the visual feature that it affects, rather than on the whole image $\mathbf{I}$. At each iteration $k$, we will apply
\begin{equation}
{\mathbf{p}}_k = {\mathbf{p}} \left( {\mathbf{x}}^P_k, {\mathbf{x}}^{P*} \right)
\text{ or }
{\mathbf{t}}_k = {\mathbf{t}} \left( \tilde{\mathbf{I}}_k, \tilde{\mathbf{I}}^* \right),
\label{eq:actionState}
\end{equation}
depending on the pre-selected (\underline{\textit{Hypothesis 3}}) action type (push or tap). In~(\ref{eq:actionState}), $\mathbf{x}^{P*}$ and $\tilde{\mathbf{I}}^*$ respectively indicate feature $\mathbf{x}^P$ and $\tilde{\mathbf{I}}$, extracted from the desired image $\mathbf{I}^*$. Since the visual features are related to the image $\mathbf{I}$, the dependency of ${\mathbf{a}}$ from $\mathbf{I}$ in~(\ref{eq:actionImg}) is maintained. Note however that since the relationship between features and image is not bijective (e.g., different images may have the same shape's outer contour), different images may lead to the same action.

The design of~(\ref{eq:actionState}) is inspired by classic IBVS, where a feature vector ${\mathbf{x}}_k$ is regulated to ${\mathbf{x}}^{*}$ by applying action
\begin{equation}
{\mathbf{a}}_k = \lambda {\mathbf{L}}_k^\dagger \left( {\mathbf{x}}^{*} - {\mathbf{x}}_k \right)
\label{eq:ibvsControl}
\end{equation}
with $\lambda$ a positive scalar gain and ${\mathbf{L}}_k$ the interaction matrix, \textit{i.e.}, the matrix such that: ${\mathbf{x}}_{k+1} - {\mathbf{x}}_k = {\mathbf{L}}_k {\mathbf{a}}_k$.

Even if we apply~(\ref{eq:ibvsControl}), there is no guarantee that $e_k$ will decrease as indicated in~(\ref{eq:errorDecrease}). First, the relationship ${\mathbf{L}}$ between dynamics of the feature ${\mathbf{x}}$ and the action ${\mathbf{a}}$ must be known and invertible, as in~(\ref{eq:ibvsControl}). Second, even if this was the case,~(\ref{eq:ibvsControl}) would only guarantee convergence of feature ${\mathbf{x}}$ to ${\mathbf{x}}^*$, which by no means implies convergence of $\mathbf{I}$ to $\mathbf{I}^*$ (since the mapping between $\mathbf{x}$ and $\mathbf{I}$ is not bijective). 

Therefore, model-based approach~(\ref{eq:ibvsControl}) seems inappropriate for defining~(\ref{eq:actionState}) to solve our problem. In the rest of this section, we examine various alternative design choices for defining~(\ref{eq:actionState}) for both tapping and pushing actions, without knowledge of the interaction matrix ${\mathbf{L}}$. In the experimental section, we will compare these design choices through the evolution of the mutual information when the actions are applied sequentially to realize a desired shape $\mathbf{I}^*$.

\subsection{Tapping}
\label{subsec:methtap}

The design of the tapping action is simpler than that of the pushing action. The reason is that we have assumed (see Sect.~\ref{sec:assumptions}) that each tapping action ${\mathbf{t}}$ levels a well-defined portion of the kinetic sand surface, without affecting the other areas. More specifically, since we rescale ${\mathbf{I}}$ to a smaller $\tilde{\mathbf{I}}$, with pixels having the same size $w_{TCP} \times h_{TCP}$ as the tool, each ${\mathbf{t}}$ should change only one pixel of $\tilde{\mathbf{I}}$. 

Since by \underline{\textit{Hypothesis 2}} we assume that the shape is feasible and that we can only lower (not raise) the kinetic sand level with the tool, the height of the desired shape is lower or equal than that of the current shape. Then, at each pixel, the difference between the image luminosities should be a monotonically increasing function of the difference between the shape heights.

Therefore, we decide to apply the tapping action on the spot (in resampled image $\tilde{\mathbf{I}}$) where the image difference is the highest. This breaks down to writing~(\ref{eq:actionState}) as: 
\begin{equation}
\begin{split}
		{\mathbf{t}} \left( \tilde{\mathbf{I}}_k, \tilde{\mathbf{I}}^* \right)  &=
		\left[
		\begin{array}{c}
			u_T \\ 
			v_T 
		\end{array}
		\right]\\ 
		& = \left[
		\begin{array}{cc}
		w_{TCP} & 0\\
		0 & h_{TCP}
		\end{array}
		\right]
		\underset{\left(u,v\right) \in \tilde{\mathbf{I}}_k}{\mathrm{argmax}}\left\| \tilde{\mathbf{I}}_k - \tilde{\mathbf{I}}^*\right\|.
		\label{eq:toolActionTappingMax}
\end{split}
\end{equation}
If $\left\| \tilde{\mathbf{I}}_k - \tilde{\mathbf{I}}^*\right\|$ has multiple maxima, we randomly choose one. Eventually -- at the following iterations -- the other maxima will be selected.

	\begin{figure*}[t]
		\centering
		\includegraphics[width=\textwidth]{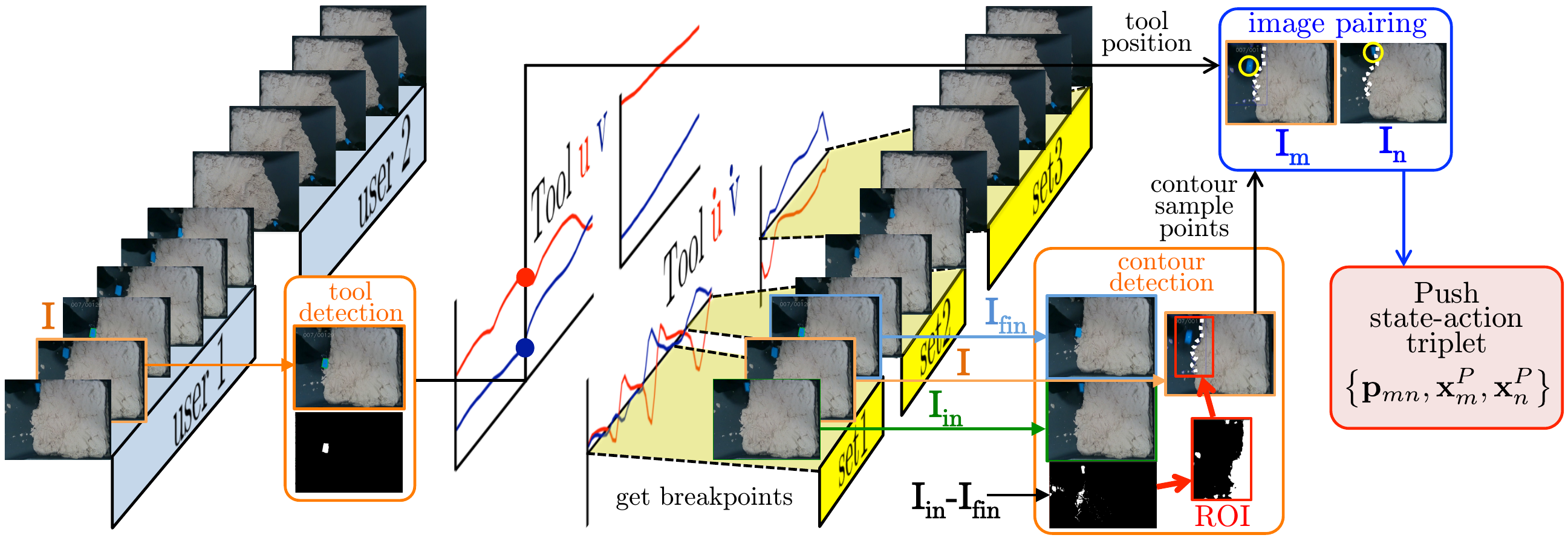}
		\caption{Left to right: image processing steps for extracting state-action triplets from the PT-Dataset. The ``pushing'' images are loaded sequentially for each user. On each image, we compute the tool position and velocity. At changes in the tool velocity, we break the sequence into smaller sets. Each set corresponds to a clear tool motion (either pushing the kinetic sand or retiring from it). Using the initial and final images of these sets, we detect the kinetic sand contour on each image. Finally, we form state-action triplets, that are characterized by a sufficiently large change in both the contours and tool positions.}
		\label{Fig:imgProc}
	\end{figure*}

\subsection{Pushing}
\label{subsec:methpush}

In contrast with tapping, in the case of pushing, the affected contours do not always have the same size and the state-action mapping is not as clear as for tapping. This has motivated us to devise a more complex strategy, which was based on the human data recorded in the \textit{PT-Dataset}. In the following, we explain the various steps of this strategy.

\subsubsection{Extracting ground truth state-action triplets from the PT-Dataset}
\label{sect:gettingTriplets}

The first step in designing the pushing actions from~(\ref{eq:actionState}) is to represent the relationship where a push action ${\mathbf{p}}_{mn}$ causes the kinetic sand contour shape ${\mathbf{x}}^P_m$ to become ${\mathbf{x}}^P_n$. For this, we define \textit{push state-action triplets} of the form:
\begin{equation}
\left\{ {\mathbf{p}}_{mn}, {\mathbf{x}}^P_m, {\mathbf{x}}^P_n \right\}.
\label{eq:triplets}
\end{equation}

We extract these triplets from the \textit{PT-Dataset} images (see Sect.~\ref{Ref:userDis}). We have two motivations in using these triplets. First, they can help design a local control strategy -- inspired by humans -- which we will explain in Section~\ref{sect.localShaping}. Second, we can use them to train a neural network to estimate -- at each iteration $m$ -- the pushing action ${\mathbf{p}}_{mn}$ required to change the kinetic sand contour from ${\mathbf{x}}^P_m$ to ${\mathbf{x}}^P_n$. 

To extract the triplets~(\ref{eq:triplets}) from the \textit{PT-Dataset}, we look for all the kinetic sand contours that have been obtained after a ``sufficiently large'' change in both contour and tool positions. Between images $\mathbf{I}_m$ and $\mathbf{I}_n$, we define the change in contour as $\left\| {\mathbf{x}}^P_{n}-{\mathbf{x}}^P_{m}\right\|$ and the change in tool position as
$\sqrt{\left(u_n-u_m\right)^2+\left(v_n-v_m\right)^2}$, with $\left( u_{i} , v_{i} \right) \in \left[0, w\right] \times \left[0, h\right]$ the image coordinates of the tool centroid in image $\mathbf{I}_i$, $i = \left\{m,n\right\}$. Therefore, we take all state-action triplets such that:
\begin{equation}
\left\{\begin{array}{l}
n > m\\
\sqrt{\left(u_n-u_m\right)^2+\left(v_n-v_m\right)^2} > \tau_u \\
\left\| {\mathbf{x}}^P_{n}-{\mathbf{x}}^P_{m}\right\| > \tau_x.
\end{array}
\right.
\label{eq:desiredContour}
\end{equation}
where $\tau_u$ and $\tau_x$ are constant, hand-tuned thresholds. Higher values of these thresholds will make criterion~(\ref{eq:desiredContour}) more selective, requiring bigger changes between the two images. 

Let us now detail the image processing pipeline that we use to extract triplets~(\ref{eq:triplets}) from the \textit{PT-Dataset} images. We process images in the HSV space since it facilitates object segmentation (the tool tip is blue, the users wear black gloves and black long sleeved shirts, the kinetic sand is light brown and the sandbox is black). Among the 14,900 images ``pushing'' images in the \textit{PT-Dataset}, we choose 3,976 (from four users) where the action and its effect on the kinetic sand are clear. 

The pipeline, shown in Fig.~\ref{Fig:imgProc}, is as follows (left to right):
\begin{enumerate}
	\item 
	The 3,976 images are loaded sequentially for each of the users. 
	\item
	For each image ${\mathbf{I}}$, we perform a \textit{tool detection}, that yields the tool position as the centroid of a blue blob, segmented in the HSV space. We discard images where the tool is not detected, and output \textit{tool position} $\left( u , v \right)$  for all other images. 
	\item
	After processing all images, the tool positions are first smoothened (we apply a weighed centered average filter of size $3$ and weights $\left[1 \quad 2  \quad 1\right]$) and then differentiated to obtain the \textit{tool velocity} on image ${\mathbf{I}}$: $\left( \dot{u} , \dot{v} \right)$.
	
	\item
	The original image sequence is reduced and broken into smaller sets, each corresponding to a clear tool motion. This is done by detecting images where the tool has either stopped (velocity norm $\sqrt{\dot{u}^2 + \dot{v}^2} < 1$ pixel/image) or changed direction (negative scalar product between consecutive velocities). We use these images as \textit{breakpoints} to split the sequence into smaller sets. Within each of these sets, the tool velocity does not change direction. We also discard images where the tool has stopped. At the end of this step, we obtain 139 sets of images. 
	
	\item
	Each of these 139 sets is processed by a \textit{contour detection} algorithm. First, we subtract final (${\mathbf{I}}_{fin}$) from initial (${\mathbf{I}}_{in}$) image in each set, to generate a difference image. In the luminosity channel of this image, lighter pixels correspond to higher differences. Then, we segment the largest blob of gray pixels in this image, and find its enclosing rectangular bounding box. This defines a ROI (Region of Interest) wherein the kinetic sand configuration has changed the most. Within this ROI, we detect \textit{on each image} ${\mathbf{I}}$ of the set the kinetic sand contour. This is done via the border following algorithm proposed by~\cite{suzuki1985topological}, and implemented in the OpenCV {\tt{findContours}} function. We then sample the kinetic sand contour with constant $N$ (here, $N = 10$), to obtain $\mathbf{x}^P$ as in~(\ref{eq:pushState}). We discard images where the contour is not detected.
	\item
	At this stage, we have obtained 2,749 sample images ${\mathbf{I}}$ distributed over 139 sets. The number has diminished from the original 3,976 since we have removed all images where the contour is not detected, or where the tool is not moving. For each ${\mathbf{I}}$, we now have the tool position $\left( u \; v \right)$ (from step 2) and contour $\mathbf{x}^P$ (from step 5). Each set is now explored to find all pairs (2-combinations) of images $\left( {\mathbf{I}}_{m}, {\mathbf{I}}_n \right)$ with sufficient contour and tool change, \textit{i.e}, pairs that comply with~(\ref{eq:desiredContour}). To this end, we must first match the sample points in ${\mathbf{x}}^P_n$ to those in ${\mathbf{x}}^P_m$; we do this by  reordering the points in ${\mathbf{x}}^P_n$, so that the sum of distances between matched pixel pairs (in ${\mathbf{x}}^P_m$ and ${\mathbf{x}}^P_n$) is minimal. After this, we can check if pair $\left( {\mathbf{I}}_{m}, {\mathbf{I}}_n \right)$ complies with~(\ref{eq:desiredContour}). The output of this last step is the set ${\cal{T}}$ of triplets of the form~(\ref{eq:triplets}). As mentioned above, each of the 139 sets corresponds to a clear tool motion, roughly 50\% of which are pushing actions and the rest `retiring' actions (the user moves the tool away from the kinetic sand between one push and the next). By checking the contour change in~(\ref{eq:desiredContour}), we discard the images of all these `retiring' actions. By checking the tool position change on all combinations of pairs ${\mathbf{I}}_{m}$, ${\mathbf{I}}_{n}$ ($n>m$) in~(\ref{eq:desiredContour}) we augment the data: from a single human pushing action between ${\mathbf{I}}_{in}$ and ${\mathbf{I}}_{fin}$ (\textit{i.e.}, one triplet), we can generate many consistent pushes. Ideally (\textit{i.e.}, if the three conditions~(\ref{eq:desiredContour}) are always met) for a set containing $N_p$ images, we will generate $\frac{N_p!}{2! \left(N_p-2\right)!}$, instead of just 1 triplet. For instance from $N_p=50$ images, we can generate 1,225 triplets. On the other hand, this is an ideal case, since consecutive images generally are too similar to meet~(\ref{eq:desiredContour}). Using $\tau_u = 5$ and $\tau_x = 3$ pixels in~(\ref{eq:desiredContour}), from the original 139 sets, we obtain ${\cal{T}}$, with $\dim\left({\cal{T}}\right)=$ 3,539 triplets.
\end{enumerate}

Finally, we derive two important statistical metrics from the set of triplets ${\cal{T}}$. These are useful to characterize the users' actions, and therefore design the local shaping strategy that will be outlined in Sect.~\ref{sect.localShaping}. 

The first metric is the mean of the distance between pairs of pixels matched on all contours. For a pair of contours $\mathbf{x}^P_m$ and $\mathbf{x}^P_n$, the distance is:
\begin{equation}
d\left( \mathbf{x}^P_m, \mathbf{x}^P_n \right) = \frac{1}{N} \sum^N_{i=1} \sqrt{\left(u_{i,n}-u_{i,m}\right)^2+\left(v_{i,n}-v_{i,m}\right)^2}.
\end{equation}
Considering all pairs $\left\{\mathbf{x}^P_m, \mathbf{x}^P_n\right\} \in {\cal{T}}$, its mean is: 
\begin{equation}
\mu\left(d\right) = \frac{1}{\dim\left({\cal{T}}\right)} \sum_{\mathbf{x}^P_m, \mathbf{x}^P_n \in {\cal{T}}} d\left( \mathbf{x}^P_m, \mathbf{x}^P_n \right).
\end{equation}
We obtain $\mu\left(d\right) = 42$ pixels, with standard deviation $\sigma\left(d\right) = 10$ pixels.

The second metric is the height (dimension along the image $v$-axis) of the smallest bounding box enclosing both contours $\mathbf{x}^P_m$ and $\mathbf{x}^P_n$, averaged over all contours in ${\cal{T}}$. For a pair of contours $\mathbf{x}^P_m$ and $\mathbf{x}^P_n$, this height is:
\begin{equation}
\begin{array}{ccc}
\Delta v \left( \mathbf{x}^P_m, \mathbf{x}^P_n \right) &=& \max \{v_{m,1} \dots, v_{m,N}, v_{n,1}\dots, v_{n,N}\} - \\ && - \min \{v_{m,1} \dots, v_{m,N}, v_{n,1}\dots, v_{n,N}\}.
\end{array}
\end{equation}
Considering all pairs $\left\{\mathbf{x}^P_m, \mathbf{x}^P_n\right\} \in {\cal{T}}$, its mean is:
\begin{equation}
\mu\left(\Delta v\right) = \frac{1}{\dim\left({\cal{T}}\right)} \sum_{\mathbf{x}^P_m, \mathbf{x}^P_n \in {\cal{T}}} \Delta v\left( \mathbf{x}^P_m, \mathbf{x}^P_n \right).
\end{equation}
We obtain $\mu\left(\Delta v \right) = 100$ pixels, with standard deviation $\sigma\left(\Delta v \right) = 22$ pixels.

\subsubsection{From global to local shaping:}
\label{sect.localShaping}
	
As mentioned above, we restrict the pushing actions 
to be along a line segment in the image plane. Although this choice limits the action space, there are still many possible choices as to where on the kinetic sand to start acting. An algorithmic approach is needed to choose the parts of the contour where to push.


Since plastic deformations of the kinetic sand are local, we opt for a local strategy. Instead of defining the push action ${\mathbf{p}}_k$ as a function of the desired contour ${\mathbf{x}}^{P*}$ (\textit{i.e.}, a function of the desired image $\mathbf{I}^*$) as indicated in~(\ref{eq:actionState}), we focus on local areas of the image, by reducing the workspace along both the $u$ and $v$ axes of the image, to define an alternative desired contour ${\mathbf{x}}^{P\star}$. 

\begin{figure}[t!]
	\centering
	\includegraphics[width=\columnwidth]{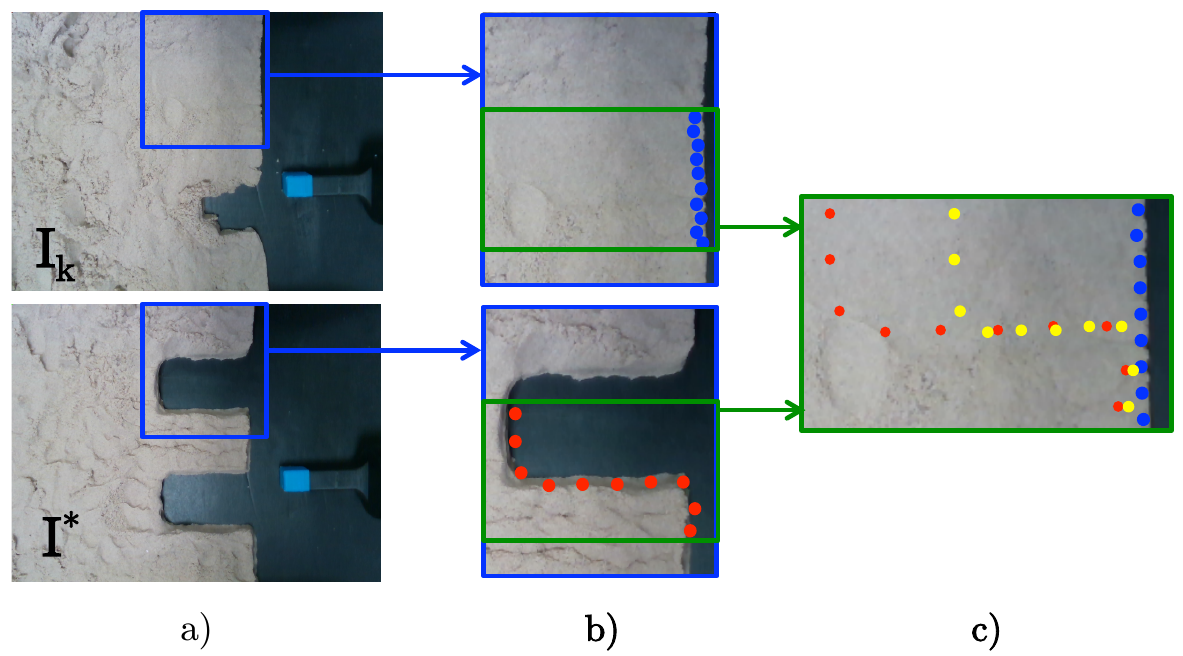}
	\caption{Local strategy for realizing the pushing action, given the current (${\mathbf{I}}_k$) and desired (${\mathbf{I}}^*$) images. a) Indentification of the rectangular ROI (blue) where the robot should act on the kinetic sand. This is obtained from the subtraction of current and desired images. b) Selection of a random ROI of lower height, resized according to the users dataset (green) and detection of sampled contours within this ROI. The contours are detected and sampled on both current (blue) and desired (red) images. c) Interpolation between current (blue) and desired (red) contours according to the users dataset, to obtain the \textit{near} contour ${\mathbf{x}}^{P\star}$ (yellow).}
	\label{fig:locStrategy}
\end{figure}

This is done through the following steps (see Fig.~\ref{fig:locStrategy}):

\begin{enumerate}
	
\item 
\textit{ROI Indentification:} This step consists in identifying a rectangular ROI (blue in Fig.~\ref{fig:locStrategy}a) where the robot should modify the most the kinetic sand. First, we subtract desired (${\mathbf{I}}^*$) from current (${\mathbf{I}}_k$) images. The following operations (blob segmentation and bounding box derivation) are identical to those presented in step 5 of the pipeline of Sect.~\ref{sect:gettingTriplets}, to identify a ROI on images in the user dataset. 
	
\item \textit{ROI Clipping:} Especially at the beginning of shape servoing, since images $\mathbf{I}$ and $\mathbf{I}^*$ (and therefore features $\mathbf{x}$ and $\mathbf{x}^*$) are most likely very different, the state error can be very large. In other words, it is unlikely that a single action would lead to the desired shape, jeopardizing convergence. Our approach is to operate in a local window by limiting the maximum ROI height (\textit{i.e.}, its dimension along the image $v$-axis). 
Clipping the ROI limits the state space -- the possible combinations of current and desired contours -- hence makes it easier to find an action. 
To determine the size of the clipped ROI, we draw inspiration from the human behavior (the \textit{PT-Dataset}). From the original ROI, we define the set of all smaller rectangular ROIs which have the same width as the original ROI, but height equal to that of the average bounding box in the dataset, $\mu\left(\Delta v \right) = 100$ (see Section~\ref{sect:gettingTriplets}). If the ROI $v$-axis size is smaller than this threshold, we do not clip the ROI. An example of clipped ROI is shown in green in Fig.~\ref{fig:locStrategy}b.

\item \textit{ROI random selection:} The natural question that arises from the previous step is which of the clipped ROIs should be selected. Our user study did not indicate any clear preference in humans as to how they choose where to apply the push actions. This means that there seems to be no clear predefined or preferred strategy as to where they start pushing, given multiple options. Moreover, in general it is safe to assume that there is more than one sequence of actions that brings the material to the desired shape. Given these observations, we decided to randomly choose the ROI within the set of clipped ROIs output at the previous step. The randomization strategy also has the benefit of getting the algorithm out of local minima. We noticed that if a fixed heuristic strategy is used here (for example always choosing the topmost among all clipped ROIs), then the same action is likely to be repeated indefinitely, without inducing much change in the kinetic sand shape. ROI randomization also helps the robot exit situations where some actions do not induce progress, e.g., due to servoing or actuation errors.

\item
\textit{Contour detection:} The selected clipped ROI is applied to both the current image ${\mathbf{I}}_k$ and the desired image ${\mathbf{I}}^*$, to detect -- once again using the OpenCV {\tt{findContours}} function -- the kinetic sand contour in both images: ${\mathbf{x}}^P_k$ and ${\mathbf{x}}^{P*}$ (with the same number of sample points, $N=10$ as used for processing the dataset). These contours are respectively blue and red in Figure~\ref{fig:locStrategy}.

\item \textit{Contour interpolation:} Although the ROI has been reduced by clipping, the difference in contours can still be large, especially at the start of shape servoing. This makes it hard or impossible to reach the desired kinetic sand shape with a single push. We therefore scale the distances between contour samples by interpolating between the current and desired contours. If the distance between ${\mathbf{x}}_k$ and ${\mathbf{x}}^*$ is higher than the dataset average distance  $\mu\left(d\right) = 42$ pixels (see Sect.~\ref{sect:gettingTriplets}), we scale it to obtain the \textit{near} desired kinetic sand contour:
\begin{equation}
{\mathbf{x}}^{P\star} = \left\{
\begin{array}{l}
{\mathbf{x}}^{P*} \hspace{2.1cm} \text{if } d\left( \mathbf{x}^P_k, \mathbf{x}^{P*} \right) \leq \mu\left(d\right),\\
{\mathbf{x}}^P_k + \frac{\mu\left(d\right)}{d\left( \mathbf{x}^P_k, \mathbf{x}^{P*} \right)} \left({\mathbf{x}}^{P*} - {\mathbf{x}}^{P}_k \right)\quad \text{otherwise. }
\end{array}
\right.
\end{equation}
Contour ${\mathbf{x}}^{P\star}$ is shown in yellow in Fig.~\ref{fig:locStrategy}c. In our work, 
there is no easy way for the robot to bring the kinetic sand back if the applied push is deeper than intended. A shorter push has the advantage of reducing the uncertainty on the successive kinetic sand shape after the push is applied. This design choice, however, comes at the cost of execution time: the robot will need to apply more pushing actions since the maximum push distance is limited.

\end{enumerate}

\begin{figure}[t!]
	\centering
	\includegraphics[width=\columnwidth]{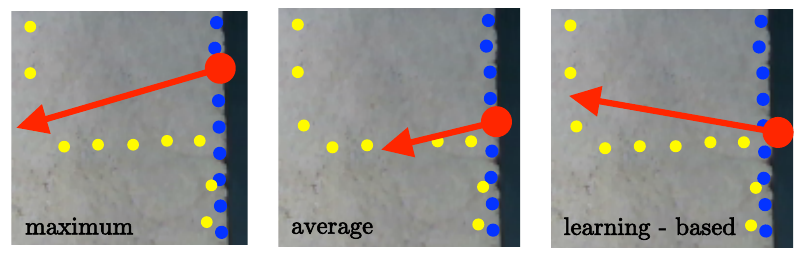}
	\caption{Alternative strategies for defining the push action (red arrow) given the current (blue) and near (yellow) contours. Left to right: maximum, average, and learning-based strategies.}
	\label{fig:pushingMethods}
\end{figure}

The interpolated contours ${\mathbf{x}}^{P\star}$ that are extracted from the clipped and randomized ROIs are used as input to the three different pushing strategies explained in the next section.

\begin{figure*}[ht!]
	\centering
	\includegraphics[width=\textwidth]{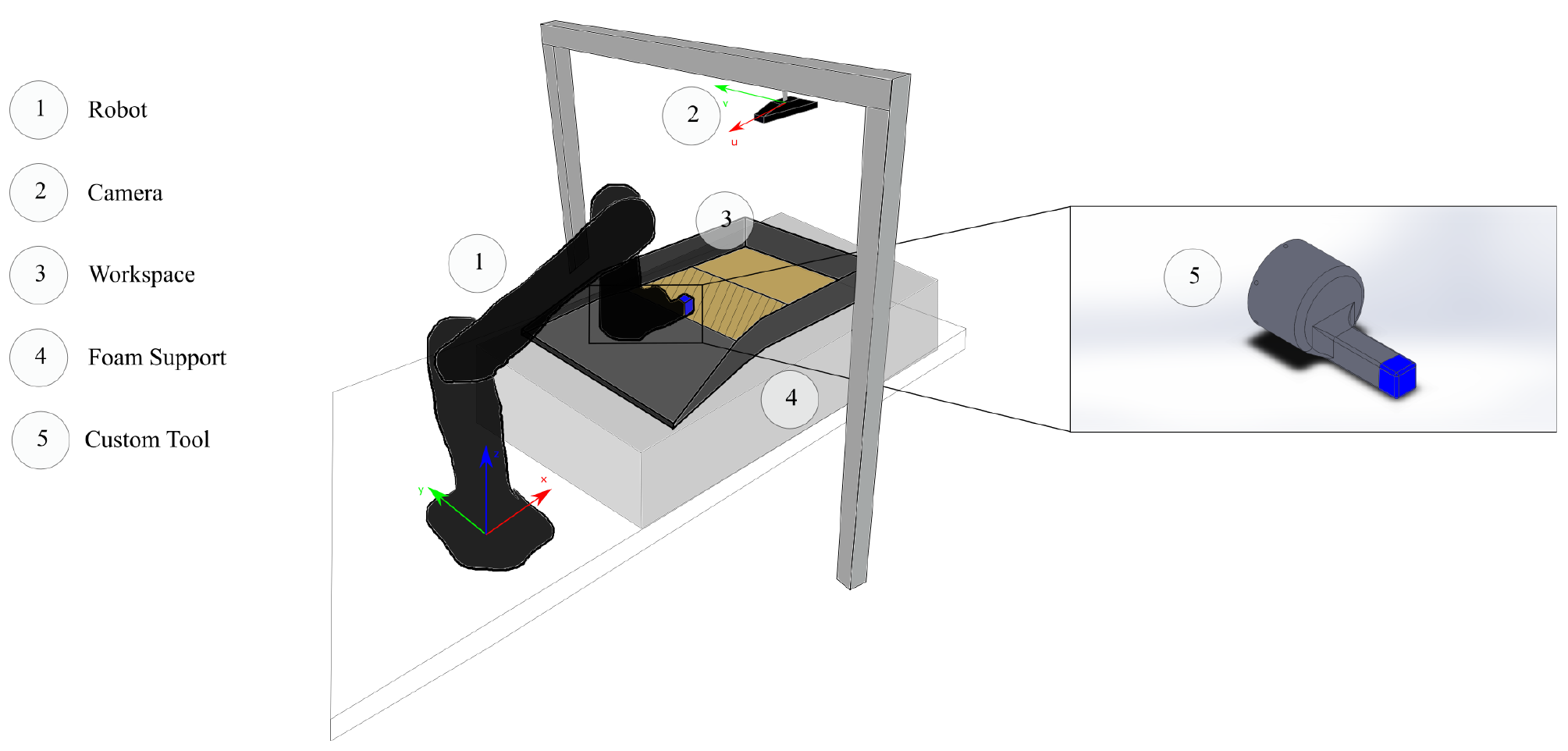}
	\caption{Experimental setup. 1) Kinova MICO robotic arm. 2) Intel RealSense camera looking downward at the workspace. 3) Workspace: a black sandbox containing the kinetic sand to be shaped. Considering its kinematics and workspace, the robot can access and manipulate only a portion, the hatched area. 
		4) Foam support that makes the sandbox comply and not break in case of abrupt robot motion. 
		5) The custom 3D-printed tool, designed to perform both pushing and tapping.}
	\label{fig:expSetup}
\end{figure*}

\subsubsection{Strategies}
\label{sect:strategies}

We devise three methods to design the pushing action based on ${\mathbf{x}}^{P}_k$ and ${\mathbf{x}}^{P\star}$:
\begin{equation}
{\mathbf{p}}_k = {\mathbf{p}} \left( {\mathbf{x}}^P_k, {\mathbf{x}}^{P\star} \right).
\label{eq:mapping}
\end{equation}
The three methods (shown in Fig.~\ref{fig:pushingMethods}) are:

\begin{enumerate}
\item \textbf{Maximum:} Given the sampled contours ${\mathbf{x}}^{P}_k$ and ${\mathbf{x}}^{P\star}$, the pushing action starts and ends at the pair of matched pixels that are the farthest (have highest Euclidean distance, among all matched pairs), on the two contours. In a nutshell, the maximum method pushes in the location where the current and desired sampled contours are the farthest. More formally, the push action is defined as
\begin{equation}
\begin{array}{l}
{\mathbf{p}}_k \left( {\mathbf{x}}^P_k, {\mathbf{x}}^{P\star} \right)  = \left[ \begin{array}{l}
u_{j,k} \\
v_{j,k} \\
u_{j,\star} \\
v_{j,\star} \\
\end{array}
\right] \text{ such that }
\\
j = \underset{i = 1 ... N}{\mathrm{argmax}}
\sqrt{\left(u_{i,k}-u_{i,\star}\right)^2+\left(v_{i,k}-v_{i,\star}\right)^2}
\end{array}
\label{eq:toolActionPushMax}
\end{equation}
If $\sqrt{\left(u_{i,k}-u_{i,\star}\right)^2+\left(v_{i,k}-v_{i,\star}\right)^2}$ has multiple maxima, we randomly choose one.

\item \textbf{Average:} While the previous method is greedy and aims at acting on the farthest contour points, with the Average method the robot pushes along the contours' centroids. Thus, the push action is defined as

\begin{equation}
{\mathbf{p}}_k \left( {\mathbf{x}}^P_k, {\mathbf{x}}^{P\star} \right) = 
\frac{1}{N} \sum^N_{i=1} \left[ 
\begin{array}{c}
u_{i,k} \\
v_{i,k} \\
u_{i,\star} \\
v_{i,\star} \\
\end{array}
\right].
\label{eq:toolActionPushAvg}
\end{equation}

\item \textbf{Learning-based:} We trained an Artificial Neural Network (ANN) using the 3,539 triplets in ${\cal{T}}$ (see Sect.~\ref{sect:gettingTriplets}). This ANN learns the mapping in~(\ref{eq:mapping}) from current and desired states to the pushing action to perform:
\begin{equation}
{\mathbf{p}}_k = {\mathbf{p}}_{\mbox{ANN}}(\mathbf{x}^P_k, \mathbf{x}^{P\star}).
\label{eq:toolActionPushANN}
\end{equation}

The input to the network is the current contour ${\mathbf{x}}^{P}_k$ and the target contour ${\mathbf{x}}^{P\star}$. With $N=10$ contour samples, the input consists of 40 scalars (the image coordinates of the sample pixels on the two contours). The output of the network is the action $\mathbf{p}_k$, which consists of 4 scalars, \textit{i.e.}, the pixel coordinates of the start and end pixels, see~(\ref{eq:toolActionPushing}).

The network architecture consists of 3 fully connected layers with 100 hidden nodes each. Early experiments conducted to choose the network architecture showed small differences. However, we found that on our particular dataset, deeper networks quickly resulted in diminishing returns (and eventually overfitting). Thus, we chose a network with sufficient capacity to perform well, but not overfit. The standard ReLu activation is used for all layers, since it is computationally efficient while modeling non-linearities (see~\cite{glorot2011deep}). We randomly split the set ${\cal{T}}$ of 3539 sample triplets into training ($\%80$), test ($\%10$) and validation ($\%10$) sets. The network is trained for 25,000 episodes where loss has converged. The training takes about fifteen minutes, and the mean absolute testing errors on the output variables are shown in Table~\ref{table:neural_network_errors}.

As shown in the table, the testing error for all output variables is within reasonable bounds, considering that the tool has a size of $30 \times 40$ pixels. Note that the trained network predicts the tool end position much better than the tool start position. 
This is a very interesting result. Clearly, the effect of a push on the contour is much more dependent on the tool end position (when it is in contact with the contour) than on its start position. The results show that the ANN has inferred this characteristic of the pushing action from the human dataset. 

\begin{table}[h!]
	\centering
	\begin{tabular}{c|c}
		Output & Mean Error (in pixels)\\ \hline
		$u_S$    & $15.4$ \\
		$v_S$    & $12.5$ \\         
		$u_E$    & $3.1$ \\          
		$v_E$    & $1.3$ \\         
	\end{tabular}
	\caption{Mean prediction error of the neural network trained to output pushing actions ${\mathbf{p}} = \left[ u_S \; v_S \; u_E \; v_E \right]^\top$ given current and desired contours.}
	\label{table:neural_network_errors}
\end{table}

\end{enumerate}

\section{Experimental Setup}\label{sect:setup}

\subsection{Objectives}

We have run a series of experiments on a robotic manipulator to validate our methods. The robot should mold the kinetic sand into desired shapes. At each iteration $k$, the molding action is determined by comparing current and desired images of the kinetic sand shape. 

The experiments rely on the hypotheses defined in Sect.~\ref{Sect:problem}. \textit{\underline{Hypothesis 1: specialised actions}} has been used to design the whole methodology (presented in the previous sections) that the robot uses to mold the kinetic sand. The objective of the experiments is to verify this Hypothesis, i.e., to check whether designing the actions according to the kinetic sand state ((\ref{eq:toolActionTappingMax}) for tapping and either~(\ref{eq:toolActionPushMax}),~(\ref{eq:toolActionPushAvg}) or~(\ref{eq:toolActionPushANN}) for pushing) is effective in shaping the kinetic sand. To deal with \textit{\underline{Hypothesis 2}}, we use as desired shapes only those feasible by the robot: matter cannot be added nor pulled, and the kinetic sand must be in the accessible robot workspace. As for \textit{\underline{Hypothesis 3}}, a human operator selects at each iteration $k$ the action to be executed (either pushing and tapping). 
		
To evaluate the quality of molding, we use mutual information error~(\ref{eq:mutInfoError}), which we calculate via the MATLAB Toolbox by~\cite{MImatlab:07}.  The software used for the experiments is available online at:~\href{https://github.com/acosgun/sand_manipulation}{https://github.com/acosgun/sand\_manipulation}. The experiments are shown in the videos available here: \href{https://www.lirmm.fr/recherche/equipes/idh/flexbot}{https://www.lirmm.fr/recherche/equipes/idh/flexbot}.

\subsection{Workspace and material}

The experimental setup is shown in Fig.~\ref{fig:expSetup} and detailed here. In all experiments, we use a Kinova Mico arm\footnote{\href{https://www.kinovarobotics.com/en}{https://www.kinovarobotics.com/en}} with 6 degrees of freedom. The tool mounted on the robot end-effector has been custom-designed and 3D-printed, so that it can be used by the robot for both pushing and tapping. We have designed the tool so that it is as similar as possible -- considering the 3D printer constraints -- to the tools used by the humans in the pilot study (see Fig.~\ref{Fig:setup}, left). Its final shape is that of a poker with a connection to the robot. 
The Intel RealSense camera -- which points downward at the sandbox -- is the same as in the user study. We position the camera so that it can view the whole sandbox as well as the tool. The walls and bottom of the sandbox are rigid (cardboard). We place a foam support under the sandbox; this was needed to avoid breaking the sandbox in the preliminary experiments, when the software had some instabilities. Considering its kinematics and workspace, the robot can access and manipulate only a portion of the kinetic sand (highlighted in Fig.~\ref{fig:expSetup} as the hatched area). Thus, we define the desired shapes only in this area of the sandbox.

We perform all computations (image processing, control and machine learning) on the CPU (Intel i7) of a Linux based computer, and use ROS Indigo along with the official Kinova-ROS packages for robot communication and control. We use OpenCV 3.0 for image processing and PyTorch to design and run the neural network. 

\subsection{Robot control for pushing and tapping}
\label{sec:robControl}

We decided to control only the translational components of the tool operational space velocity. With reference to the robot frame shown in Fig.~\ref{fig:expSetup}, $\mathbf{v}_x$ and $\mathbf{v}_y$ (the velocity components that are parallel to the sandbox plane) are controlled using visual servoing, while $\mathbf{v}_z$ is controlled to regulate the tool height along the $z$-axis. Since the tool can be easily detected as the centroid of a blue blob, IBVS provides a robust and elegant way of regulating it in the image. Furthermore, since the image plane is parallel to the $xy$ plane, the formulation of the interaction matrix is simple and the scene depth and camera focal length can be included directly in the control gain. In a nutshell, to drive the tool centroid from the current pixel $\left(u, v\right)$ to the desired one $\left(u^*, v^*\right)$ we can simply apply:
\begin{equation}
\left[
\begin{array}{c}
\mathbf{v}_x \\
\mathbf{v}_y
\end{array}
\right] =  \overline{\mathbf{v}}_{xy} \widehat{
\left[
\begin{array}{c}
- u^* + u\\
v^* - v
\end{array}
\right]
}
\label{eq:visServo}
\end{equation}
where the minus sign on the first component is due to the orientation of image and robot frame axes (see Fig.~\ref{fig:expSetup}), and the $\hat{}$ symbol indicates error vector normalization. We introduce this to avoid asymptotic convergence (in which case the velocities become very small as the error diminishes) and to maintain the velocity norm in the $xy$ plane constantly equal to pre-tuned value $\overline{\mathbf{v}}_{xy}>0$. Simultaneously, we regulate the tool height to $z^*$ using:
\begin{equation}
\mathbf{v}_z = \overline{\mathbf{v}}_z \sign{\left( z^* - z \right)}.
\label{eq:posZ}
\end{equation}
As for $\mathbf{v}_x$ and $\mathbf{v}_y$, the $\sign$ function is introduced to avoid asymptotic convergence and to maintain the velocity norm constant and equal to pre-tuned value $\overline{\mathbf{v}}_{z}>0$.

We can apply control laws~(\ref{eq:visServo}) and~(\ref{eq:posZ}) on each acquired image, until the visual error is below some threshold, at which point the task is deemed finished.

\begin{figure}[t!]
	\centering
	\includegraphics[width=\columnwidth]{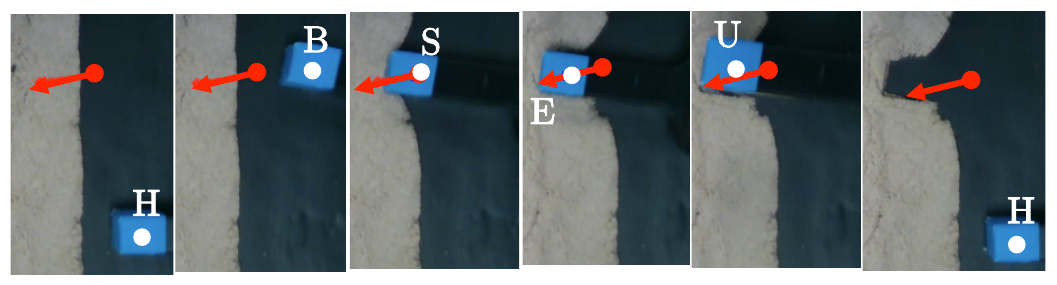}
	\caption{Steps of the IBVS control scheme for pushing the kinetic sand, here using action $\mathbf{p}$ (red). The robot tool is blue and the waypoint pixels (H, B, S, E and U) are white.}
	\label{fig:wayPoints}
\end{figure}

For tapping, the desired pixel $\left(u^*, v^*\right)$ is set to ${\mathbf{t}} = \left( u_T \; v_T \right)$ derived from~(\ref{eq:toolActionTappingMax}).
For pushing, the tool must move sequentially first to the start pixel $\left( u_S \; v_S \right)$, then to the end pixel $\left( u_E \; v_E \right)$, which are defined according to either~(\ref{eq:toolActionPushMax}),~(\ref{eq:toolActionPushAvg}) or~(\ref{eq:toolActionPushANN}). 

Yet, for both actions, waypoints are needed to guarantee that the tool does not touch the kinetic sand nor occlude the camera view when it is not intended to. Both the pushing and tapping motions start and end at a constant home pixel $H$, placed on the sandbox far from the kinetic sand. It is only when the motion is finished and the tool has returned at $H$, that the acquired image is processed to determine the next action and the iteration index $k$ is increased. Since the tool is at $H$, there is no risk of camera occlusion and the kinetic sand has stopped moving. On the other hand, the images acquired by the camera while the tool moves between the waypoints are only used to drive the IBVS control scheme according to~(\ref{eq:visServo})-(\ref{eq:posZ}), not to determine the next $\mathbf{p}$ or $\mathbf{t}$.

For pushing, these waypoints are shown in Fig.~\ref{fig:wayPoints}.  The second waypoint $B$ is placed on the same pixel row as the start pixel $S$, but on the side opposite to the kinetic sand. This waypoint is indispensable to avoid accidentally hitting the kinetic sand while moving from $H$ to $S$. After $B$, the tool moves to $S$ and then $E$ (this is the actual \textit{push action} $\mathbf{p}$). We move the tool along these waypoints, using only control law~(\ref{eq:visServo}), and setting $\mathbf{v}_z = 0$. Then, the tool is raised to a waypoint $U$ placed at higher $z$; for this, we only apply control law~(\ref{eq:posZ}) on $\mathbf{v}_z$, and set $\mathbf{v}_x = \mathbf{v}_y = 0$. Finally, the tool is brought back to the home pixel, using both~(\ref{eq:visServo}) and~(\ref{eq:posZ}). Similarly, to tap the kinetic sand, the tool must first rise from $H$ to a given height (above kinetic sand level), then translate to a waypoint above $T$ (defined by the tapping action $\mathbf{t}$), lower to $T$, rise again, and finally return to $H$. 

\begin{figure*}[ht!]
	\centering
	\begin{subfigure}[h]{0.9\textwidth}
		\centering\includegraphics[width=\textwidth]{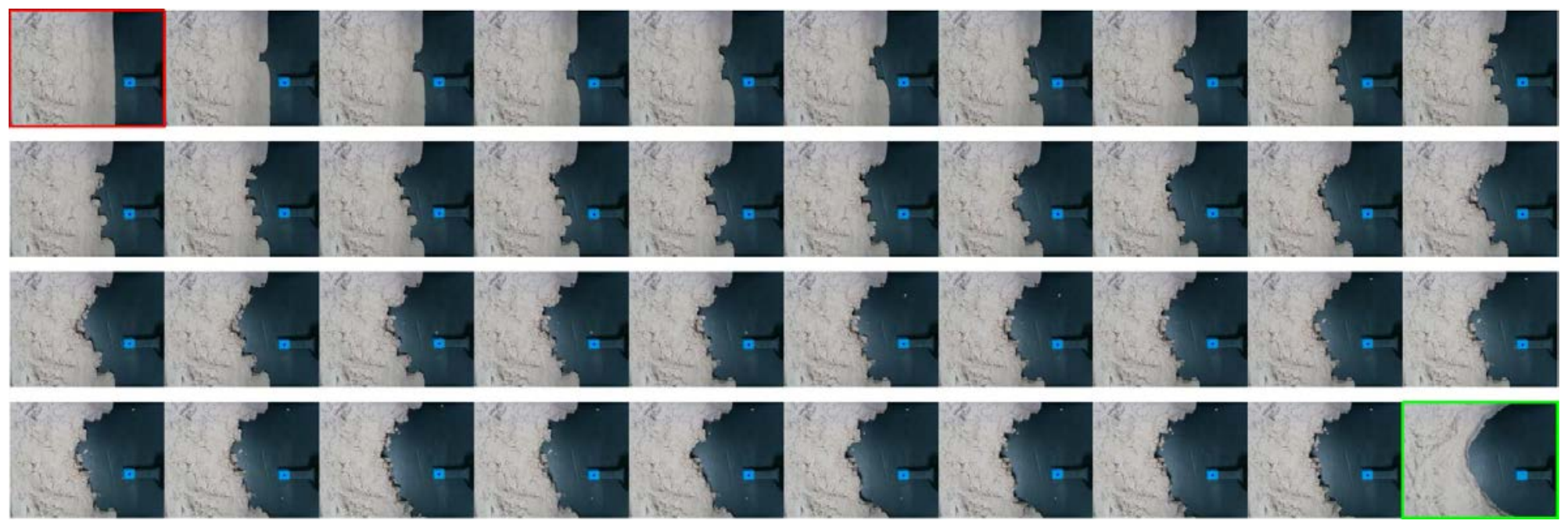}
		\caption{Desired C-shape.}
	\end{subfigure}
	\begin{subfigure}[h]{0.9\textwidth}
		\centering\includegraphics[width=\textwidth]{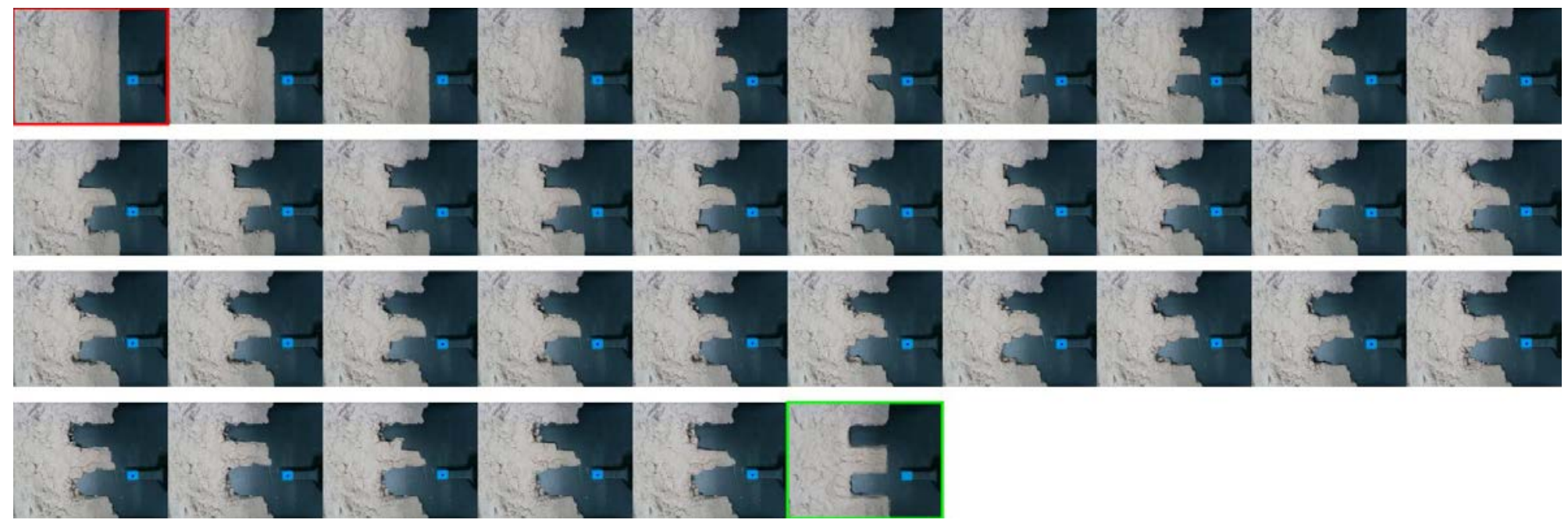}
		\caption{Desired E-shape.}
	\end{subfigure}
	\begin{subfigure}[h]{0.9\textwidth}
		\centering\includegraphics[width=\textwidth]{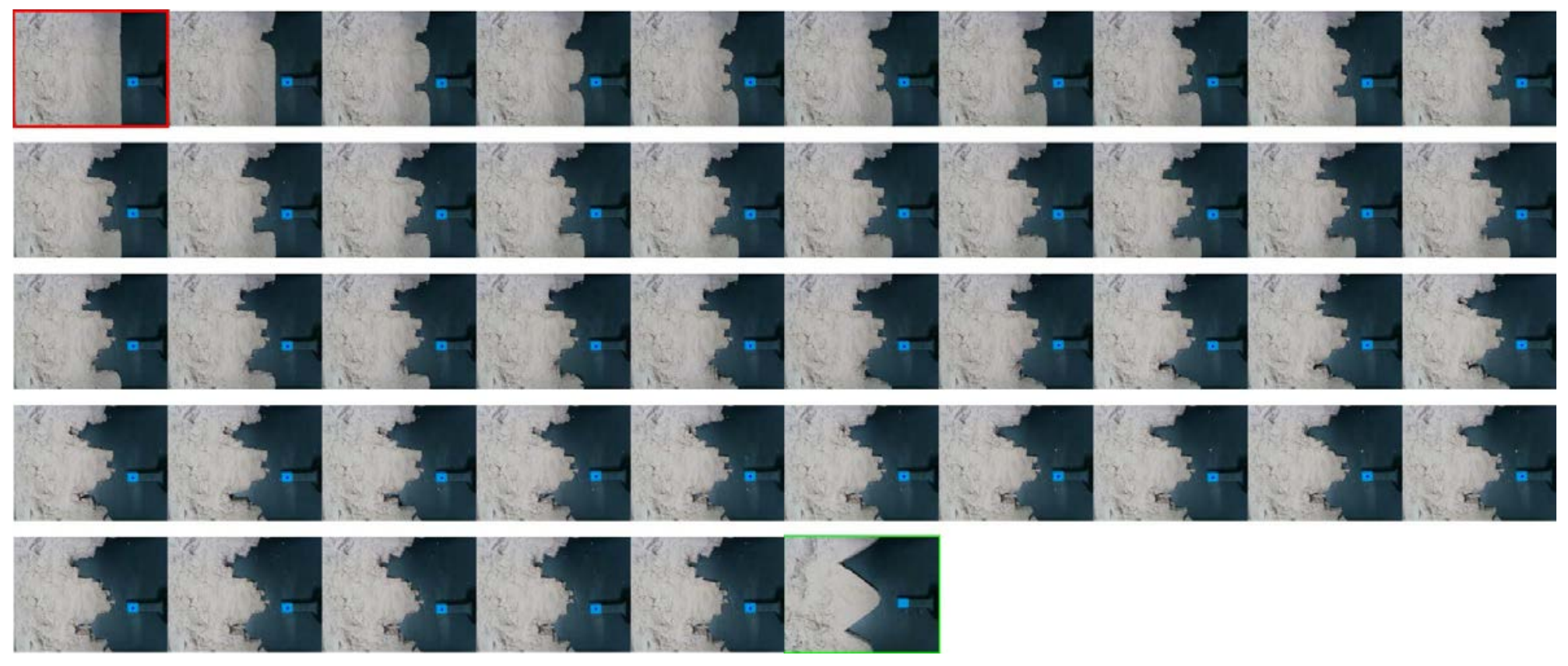}
		\caption{Desired $\Sigma$-shape.}
	\end{subfigure}
	
	\caption{Images from three experiments with only pushing actions obtained with the \textit{maximum} strategy. The three images framed in red show the initial state of the kinetic sand. The three images framed in green are the desired shapes (top to bottom: C, E and $\Sigma$). Intermediate images show the results of each consecutive pushing action.}
	\label{fig:CESexperiments}
\end{figure*}

\begin{figure*}[t]
	\centering
	\includegraphics[width=\textwidth]{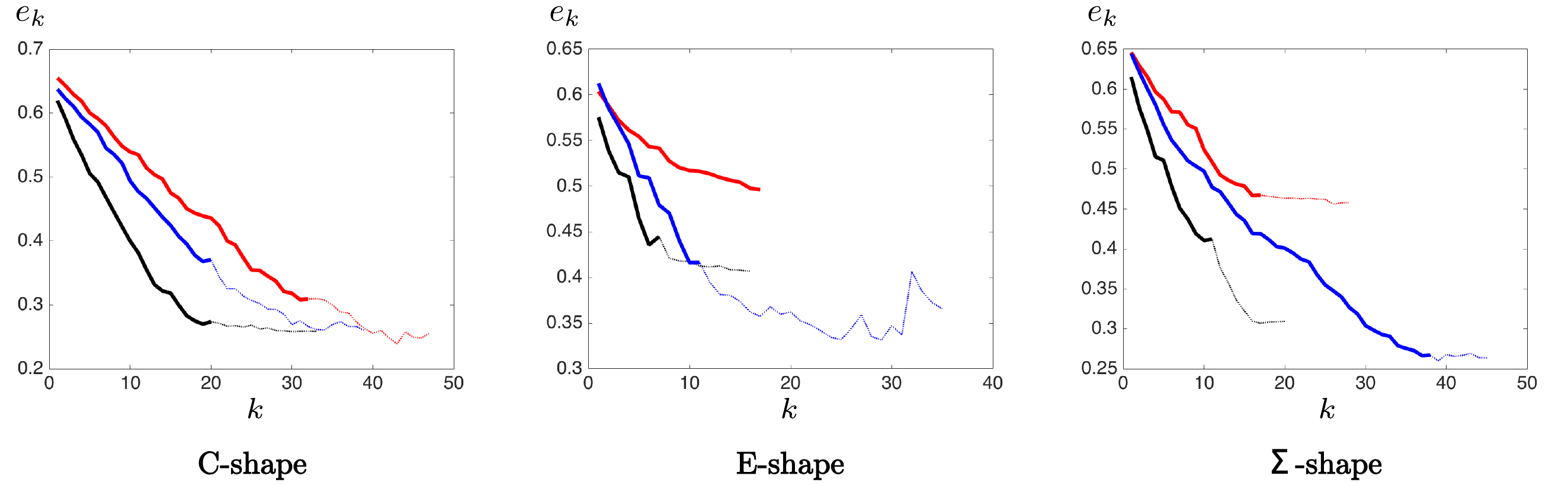}
	\caption{Evolution of the mutual information error $e_k$ at each iteration $k$ in pushing experiments with three desired shapes. The colors correspond to the three strategies: maximum (blue), average (red) and learning-based (black). The solid curves show $e_k$ until ther termination condition $e_{k+1} > e_k$, whereas the dashed curves show the error until manual termination by the human operator.}
	\label{fig:mutInfoPush}
\end{figure*}

\section{Experimental Results}\label{sect:experiments}


We have run a number of experiments to test the two actions first separately, and then together. In all cases, the human operator must press a key to launch the action at each iteration $k$. He also terminates the experiment when he considers that the performance cannot further improve. This is a subjective choice, that we would like to avoid, by finding an objective termination condition -- related to the image error $e_k$ -- for the robot to stop autonomously. In the experiments with both pushing and tapping together, the human also selects -- via the keyboard -- which action to perform at each iteration $k$, depending on the current kinetic sand state. In all the experiments, the desired shapes $\mathbf{I}^*$ are images acquired after a human has shaped the kinetic sand by hand, trying at best to guarantee \textit{\underline{Hypothesis 2}} (feasible shape).


\subsection{Pushing experiments}
\label{sect:pushXP}

The results of the pushing experiments are shown in Figures~\ref{fig:CESexperiments} and~\ref{fig:mutInfoPush}. We used three desired shapes (reminiscent of letters C, E and $\Sigma$) framed in green in Fig.~\ref{fig:CESexperiments}. For each shape, the robot has to mold the kinetic sand using the three strategies: maximum~(\ref{eq:toolActionPushMax}), average~(\ref{eq:toolActionPushAvg}) and learning-based~(\ref{eq:toolActionPushANN}). In all nine experiments, we start from an initial shape with straight contour (red framed in Fig.~\ref{fig:CESexperiments}). 

Figure~\ref{fig:CESexperiments} shows the sequence of images obtained after each pushing action, using the maximum strategy. The final shapes obtained by the robot are shown in the image preceding the green framed ones. The differences are mainly due to the tool resolution and to the fact that we decided not to control the tool orientation. 

In Fig.~\ref{fig:mutInfoPush} we plot the mutual information error $e_k$ from~(\ref{eq:mutInfoError}) at each iteration $k$, for each desired shape (left to right: C, E and $\Sigma$) and each strategy (maximum in blue, average in red and learning-based in black). This experiment is also useful to see if it is possible to identify a termination condition for the molding. Inspired by~(\ref{eq:errorDecrease}), we test the following condition: the robot must stop when the mutual information error increases, \textit{i.e.}, at iteration $k+1$ such that $e_{k+1} > e_k$. The solid curves show the values of $e_k$ until this condition is verified, whereas the dashed plots represent the error values until manual termination by the operator\footnote{Coincidentally, the two termination conditions are identical in the average E-shape experiment (red curve in the center).}. 

Let us first comment the solid curves. Note that even for the same desired shape, since the initial image is not identical in the three strategies, the initial values are slightly different. Nevertheless, for all nine plots the general trend is decreasing. We also note that the strongest slope -- at least in the initial iterations -- is obtained with the learning-based approach (black). The second best is maximum (blue), followed by average (red). However, for shapes E and $\Sigma$, the learning-based strategy is terminated earlier than the maximum one, which ends up with the lowest overall error (0.42 for E shape, and 0.27 for $\Sigma$). The reason is most probably due to the data distribution: the neural network has not been trained on small contour variations and is therefore less efficient in such situations. Among the three methods, average (red) is the worse. This could be expected, since 
this heuristic 
is easily driven into local minima (particularly for very convex shapes, as E and $\Sigma$).

Comparing the dashed and solid curves, it is noteworthy that most strategies continue to reduce error $e_k$ even after having reached the termination condition. This is probably due to the complexity of the task, which can be seen as a non-convex optimization problem. In brief, an action may occasionally increase $e_k$ (for an iteration), but on a longer time horizon the error can still diminish. This leads to questioning the convergence condition~(\ref{eq:errorDecrease}), which may be too strict for the shape servoing task.
	
	\begin{figure}[t]
		\centering
		\includegraphics[clip,trim=0.1cm 0 0.1cm 0, width=\columnwidth]{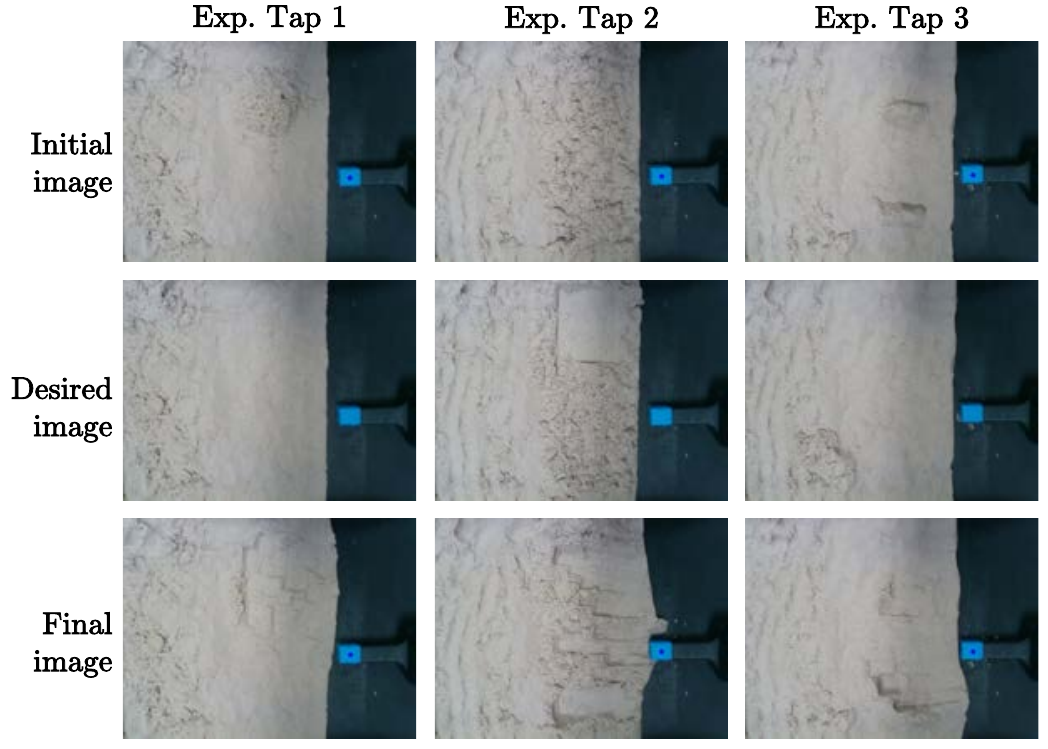}
		\caption{Three tapping experiments (left to right columns) characterized by different initial and desired images (top and middle row). The final image obtained by the robot is shown in the bottom row for each experiment.}
		\label{fig:tapExperiments}
	\end{figure}
	
\subsection{Tapping experiments}
\label{Sect.tapExp}

To assess the tapping strategy we have run three experiments, depicted in Fig.~\ref{fig:tapExperiments} from left to right. The tool size is $w_{TCP}\times h_{TCP} = 30 \times 40$  pixels. The experiments are characterized by different initial (upper row of images) and desired (middle row) conditions on the kinetic sand height, while the contours were kept unchanged. On the lower row of the figure we show the images obtained by the robot after having applied tapping strategy~(\ref{eq:toolActionTappingMax}) for a few iterations. The termination condition is given by the human operator when he esteems that there will be no further improvements (\textit{i.e.}, after respectively 30, 19 and 24 images).

\begin{figure*}[t]
	\centering
	\includegraphics[clip,trim=0.1cm 0 0.1cm 0, width=0.9\textwidth]{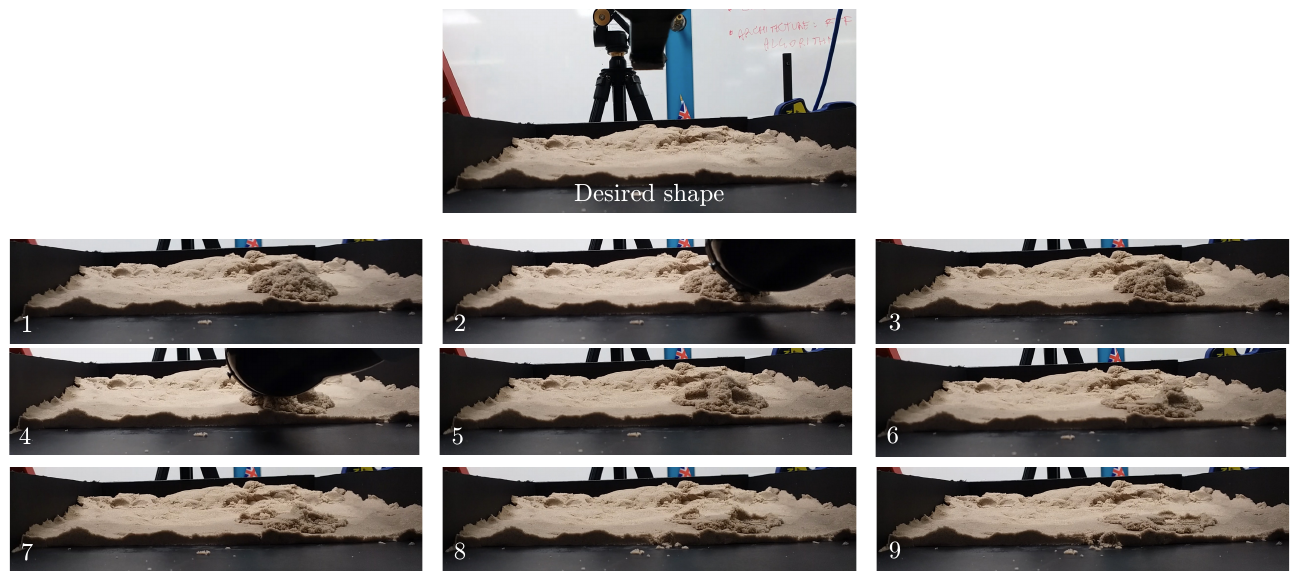}
	\caption{Side view of the first tapping experiment. Top: desired shape. Bottom: sequence of nine consecutive snapshots during the experiment (the robot arm is visible in the second and fourth snapshots). Note that the kinetic sand is leveled as required.}
	\label{fig:tapSide}
\end{figure*}

Experiments 1 and 3 have similar desired images (both require flattening the whole accessible workspace) but differ in the initial image. Experiment 1 starts with a pile of kinetic sand in the upper part of the image, whereas Experiment 3 starts with two smaller piles on each side of the workspace. Experiment 2 starts with loose kinetic sand and the desired task consists in flattening a square in the upper part of the image. As the final images show, in Experiments 1 and 3 the robot manages to flatten the piles, whereas in Experiment 2 it acts over the whole workspace and not only on the specified square. The mutual information error does not improve during any of the three experiments. In our opinion, this is due to two reasons indicated below.
\begin{itemize}
	\item 
	Even more than for pushing, because of the tool size and of the robot characteristics, the robot cannot tap on the kinetic sand as accurately as a human. Typically, since the constant tool height reference $z^*$ is not related to the -- varying but unmeasurable -- kinetic sand height, the tool often penetrates the kinetic sand, and leaves a footprint with a rectangular shade that appears in the robot images, but not in the human ones (compare bottom and middle row of images). We also tried replacing position control with force control along the $z$ axis, but the Kinova embedded force sensing is not accurate enough for this.
	\item
	Feature $\tilde{\mathbf{I}}$ is not the best feedback for this action. Indeed, since the goal of tapping is to modify the kinetic sand shape along directions perpendicular to the image plane, the best feedback would rather be a point cloud from a depth image. Yet, as we mentioned, the RealSense depth image is not exploitable. The RGB image cannot characterize directly the kinetic sand depth, because of effects such as the material granularity and shades. This confirms the formidable capacity of the human sensorimotor system, which through stereovision and experience alone, can mold very precisely along the depth axis. Despite the questionnaire results, probably haptic feedback also plays a more important role here, than it does for pushing.
\end{itemize}

Although $\tilde{\mathbf{I}}$ and generally RGB data is inappropriate for tapping, the designed action does properly level the kinetic sand. This is visible in Fig.~\ref{fig:tapSide}, where we show a side view of Experiment 1, with the desired shape (top) and a sequence of nine consecutive snapshots during the experiment (bottom). The figure clearly shows that the kinetic sand level is gradually reduced where required, and that a side view such as this one would be much more useful -- as feedback signal for tapping -- than the top view used in our work.

Also note that tapping on the kinetic sand boundaries may cause its expansion and alter the shape contours (see the bottom row of images in Fig.~\ref{fig:tapExperiments}). Nevertheless, our framework can overcome this issue by alternating pushing and tapping actions until convergence of the overall error. In the next Section, we present the results of experiments where we alternate between pushing and tapping.

\begin{figure}[t]
	\centering
	\includegraphics[clip,trim=0.1cm 0 0.1cm 0, width=\columnwidth]{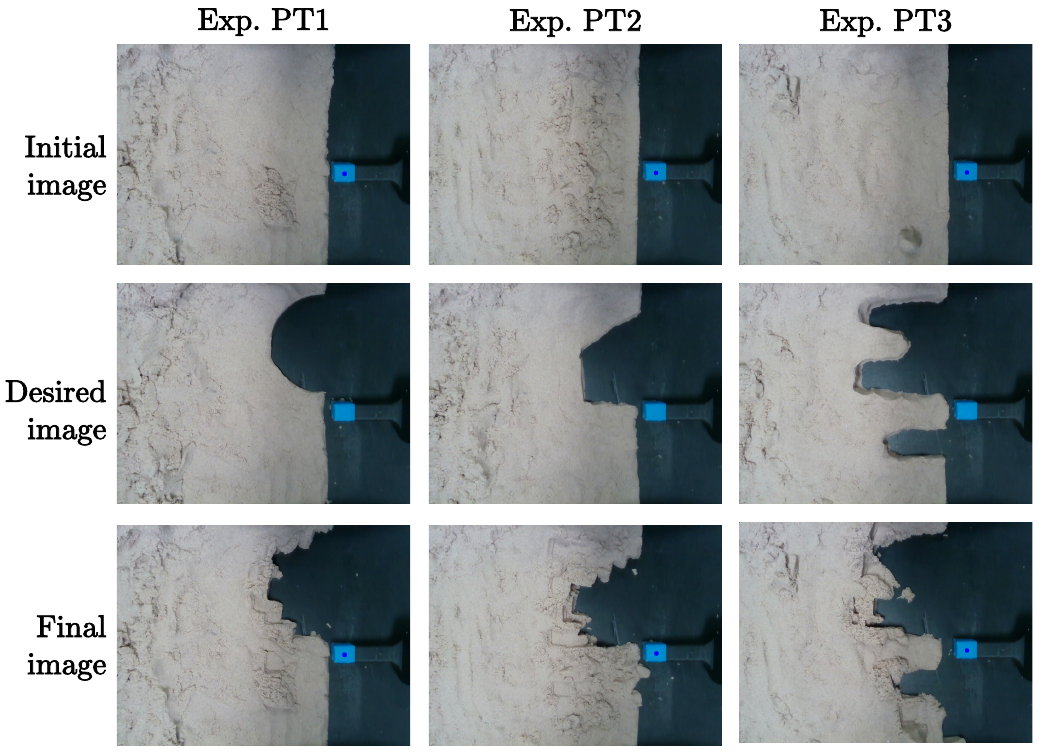}
	\caption{Three experiments requiring both pushing and tapping (left to right columns) characterized by different initial and desired images (top and middle row). The final image obtained by the robot is shown in the bottom row for each experiment.}
	\label{fig:pushtapExperiments}
\end{figure}

\subsection{Experiments requiring both pushing and tapping}
\label{Sect.pushtapExp}

Finally, we have run three experiments requiring both pushing and tapping actions, and depicted in Fig.~\ref{fig:pushtapExperiments} from left to right. Since our framework is not yet capable of autonomously selecting the best action type at each iteration (\textit{\underline{Hypothesis 3}}), the human operator chooses it using the keyboard. Among the pushing strategies, he can also choose between \textit{maximum} and \textit{learning-based}, since these performed better than \textit{average} in the experiments of Sect.~\ref{sect:pushXP}. In short, at each iteration of these three experiments, the operator can select from the keyboard between: tap action, push action using maximum strategy and push action using learning-based strategy.

The experiments are characterized by different initial (upper row of images) and desired (middle row) conditions on both the kinetic sand height and contours. On all initial images the contour is a straight line and on all desired images all the workspace kinetic sand has been flattened. The experiments differ in the desired contour and in the initial height: a wide and low pile of kinetic sand in Exp. PT1, loose kinetic sand everywhere in Exp. PT2, and a thin and high pile in Exp. PT3. 

\begin{figure}[t]
	\centering
	\includegraphics[width=\columnwidth]{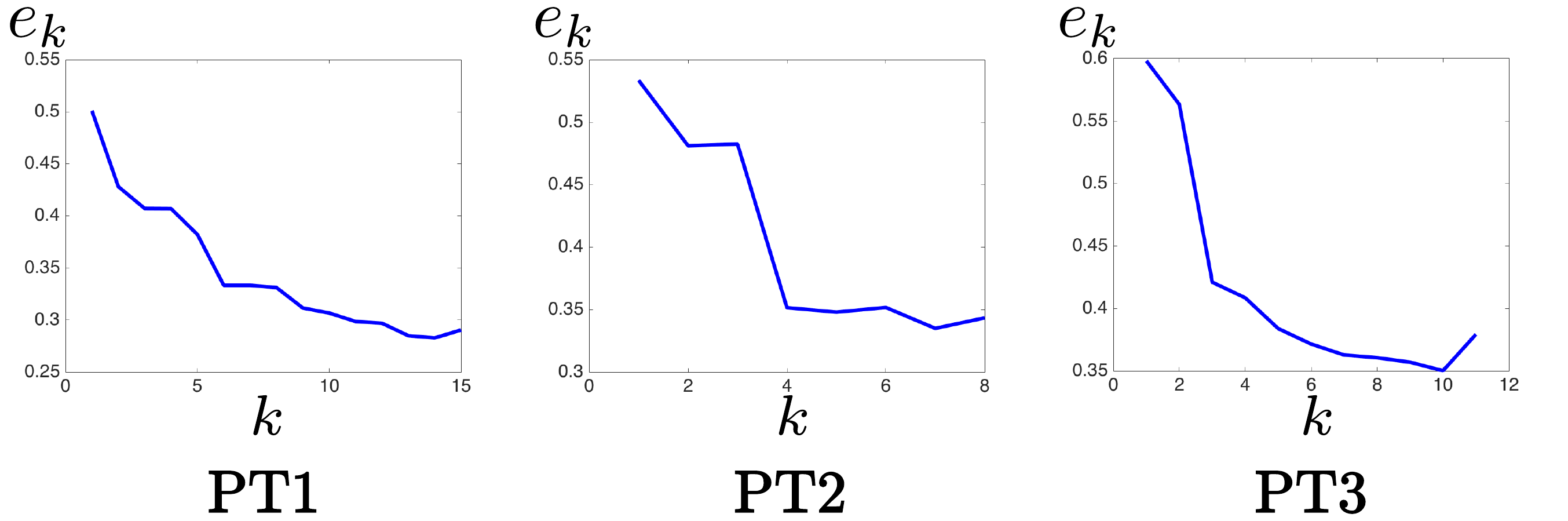}
	\caption{Evolution of the mutual information error $e_k$ at each iteration $k$ in experiments with both pushing and tapping, to obtain three desired shapes.}
	\label{fig:mutInfoPushTap}
\end{figure}

On the lower row of the figure we show the images obtained by the robot after having applied pushing and tapping actions for a few iterations. As the final images show, the three experiments confirm that pushing is more effective than tapping. In fact, we can see qualitatively that all three final contours resemble the desired ones: by pushing, the robot has even corrected the contour expansion effect of tapping, mentioned in Sect.~\ref{Sect.tapExp}. Instead, in Exp. PT2 the robot has not tapped the whole workspace as required. In Fig.~\ref{fig:mutInfoPushTap}, we have plotted $e_k$ at each iteration for the three experiments. Since the termination condition used in the pushing experiments of Sect.~\ref{sect:pushXP} seemed too conservative, we have decided to relax it and use $e_{k+1} - e_k > 0.005$. In practice, we tolerate a maximum increase of $0.005$ in mutual information, from one iteration to the next. We plot the curves until this condition met. 
The curves confirm the results seen in Fig.~\ref{fig:pushtapExperiments}: in all three cases, using the proposed termination condition, the error is reduced by more than 0.2. Furthermore, the choice of this condition is appropriate, because using $e_{k+1} > e_k$ as in Sect.~\ref{sect:pushXP} would have interrupted the robot too early on experiments PT1 and PT2, because of the weakness of the tapping action. Using $e_{k+1} > e_k$ instead of $e_{k+1} - e_k > 0.005$, the robot would have stopped: for PT1 after 7 iterations at $e_7 = 0.33$ instead of after 15 iterations at $e_{15} = 0.29$, and for PT2
after 3 iterations at $e_3 = 0.48$ instead of after 8 iterations at $e_{8} = 0.34$.

\subsection{Discussion}
\label{sect:discussion}

The experiments show two important limitations of our framework: first, it cannot regulate the tool orientation during neither pushing nor tapping; second, it is not capable of autonomous selection of the action type to be applied at each iteration. In the following paragraphs we propose solutions to these two problems. We also comment on the robustness of our framework with respect to the variability of experimental conditions.

\subsubsection{Controlling the Tool Orientation}
\label{sec:toolOrient}

As mentioned in Sect~5.1, if the contact between tool and material can be approximated by a point or by a sphere, the effect of pushing and tapping on the kinetic sand will be invariant to the tool orientation. Since the tool we use in our setup does not fulfill such hypothesis, its orientation will affect the kinetic sand shape. We hereby explain how one could add tool orientation control to our framework.

Among the three tool orientations in space, let us shortly discuss the one around the camera optical axis, which is the easiest to visually measure and control in our setup. For tapping, the definition of the controlled feature $\tilde{\mathbf{I}}$ explicitly accounts for the tool resolution and tolerates that changes in the image $\tilde{\mathbf{I}}$ cannot be smaller than the tool size. Hence, rotating the tool will not enhance the performances. For pushing, orienting the tool is more interesting since it could avoid the differences between obtained and desired images visible, for example, in Figures 9 and 13. The orientation can be included in ${\mathbf{p}}$:
\begin{equation}
{\mathbf{p}}_k = \left[ u_k \; v_k \; \theta_k \; u_\star \; v_\star \; \theta_\star \right]^\top.
\label{eq:toolActionPushing}
\end{equation}	
One way of designing $\theta_k$ and $\theta_\star$ is by aligning the tool with the translational direction, i.e., setting:
\begin{equation}
\theta_k = \theta_{\star} = \atantwo \left( u_\star - u_k, v_\star - v_k \right).
\end{equation}
Alternatively, one can design $\theta_k$ and $\theta_{\star}$ independently from the action start and end pixels. The design will differ for each of the three pushing strategies of Sect.~\ref{sect:strategies}. For the \textit{maximum} strategy, the most intuitive solution is to take as start and end orientations those of the normals to the contours ${\mathbf{x}}^P_k$ and ${\mathbf{x}}^{P\star}$, at the points where the contours are the farthest.
Naming 
\begin{equation}
n \left( u, v, {\mathbf{x}}^P \right) \in \left] -\pi, \pi \right]
\end{equation}
the orientation of the normal to contour ${\mathbf{x}}^P$ at pixel $\left( u, v \right)$, the \textit{maximum} strategy would yield:
\begin{equation}
\theta_k = n \left( u_{j,k}, v_{j,k}, {\mathbf{x}}^P_k \right) \quad \quad \theta_{\star} = n \left( u_{j,\star}, v_{j,\star}, {\mathbf{x}}^{P\star} \right)
\end{equation}
such that:
\begin{equation}
j = \underset{i = 1 ... N}{\mathrm{argmax}}
\sqrt{\left(u_{i,k}-u_{i,\star}\right)^2+\left(v_{i,k}-v_{i,\star}\right)^2}.
\label{eq:toolActionPushMax}
\end{equation}
For the \textit{average} strategy, one could take the average orientation of all $N$ normals to contours ${\mathbf{x}}^P_k$ and ${\mathbf{x}}^{P\star}$:
\begin{equation}
\begin{array}{l}
\theta_k = \frac{1}{N} \sum^N_{i=1} n \left( u_{i,k}, v_{i,k}, {\mathbf{x}}^P_k \right) \\
\theta_\star = \frac{1}{N} \sum^N_{i=1} n \left( u_{i,\star}, v_{i,\star}, {\mathbf{x}}^{P\star} \right)
\end{array}
\label{eq:toolActionPushAvg}
\end{equation}
Finally, for the \textit{learning-based} strategy, the start and end orientations should be learned by the Artificial Neural Network using the dataset. This requires extracting the tool orientation from each image in the dataset, i.e., adding such feature to the output of the image processing pipeline in Sect.~\ref{sect:gettingTriplets}.

To make the robot rotate the tool while pushing, the visual servoing controller in Sect.~\ref{sec:robControl} must also be adapted. Since the current and desired tool orientations are directly measurable in the image, this can be done, as for the components of $\mathbf{v}$, through a feedback controller on the tool angular velocity around the optical axis, $\mathbf{\omega}_z$:
\begin{equation}
\mathbf{\omega}_z = \bar{\mathbf{\omega}}_z \sign \left( \theta^* - \theta_k \right).
\end{equation}
with $\bar{\mathbf{\omega}}_z$ a pre-tuned positive scalar.

\subsubsection{Action selection}

It could be possible to automatically choose the action to be realized at each iteration $k$. One way of doing this is by verifying on the current image $\mathbf{I}_k$ which feature ``requires the most change'' to look as it does in desired image $\mathbf{I}^*$. This can be done by comparing the Euclidean distances between current and desired push- and tap-controlled features. Since these features (respectively contour $\mathbf{x}^P$ and image $\tilde{\mathbf{I}}$) are defined in different sets and expressed in different measurement units, we must include some positive weight $\alpha > 0$ to make the comparison consistent. Note that the tuning of scalar weight $\alpha$ is crucial here. Choosing a high (respectively, low) value will make tapping (respectively, pushing) prevail more often. The value of $\alpha$ could also be learned from the dataset. The described algorithm would look as follows.

\begin{algorithm}[h!]
	\vspace{0.5mm}
	\noindent\textbf{Algorithm:} Automatic selection of action type $\mathbf{a}_k$ at iteration $k$
	\vspace{0.5mm}
	\hrule
	\vspace{1.5mm}
	\begin{algorithmic}[1]
		\renewcommand{\algorithmicrequire}{\textbf{Input:}}
		\renewcommand{\algorithmicensure}{\textbf{Output:}}
		\REQUIRE Desired image $\mathbf{I}^*$ and current image $\mathbf{I}_k$.
		\ENSURE Action type $\mathbf{a}_k$ (either push $\mathbf{p}$ or tap $\mathbf{t}$).
		\hrule
		\vspace{1.5mm}    
		\STATE Extract $\mathbf{x}_k^P$ from $\mathbf{I}_k$ and $\mathbf{x}^{P*}$ from $\mathbf{I}^*$
		\STATE Resize $\mathbf{I}_k$ to $\tilde{\mathbf{I}}_k$ and $\mathbf{I}^*$ to $\tilde{\mathbf{I}}^*$
		\IF {$\left\| \mathbf{x}^P_k - \mathbf{x}^{P*} \right\| >  \alpha \left\| \tilde{\mathbf{I}}_k - \tilde{\mathbf{I}}^*\right\|$} 
		\STATE $\mathbf{a}_k \leftarrow \mathbf{p} \left( \mathbf{x}_k^P, \mathbf{x}^{P*} \right)$ \COMMENT{push the contours}
		\ELSE 
		\STATE $\mathbf{a}_k \leftarrow \mathbf{t} \left( \tilde{\mathbf{I}}_k, \tilde{\mathbf{I}}^* \right)$ \COMMENT{tap the surface}
		\ENDIF
		\RETURN $\mathbf{a}_k$
		\end{algorithmic}
		\hrule
		\hspace{1cm}
\end{algorithm}

\subsubsection{Robustness to variability of the experimental conditions}

\begin{figure}[t]
	\centering
	\includegraphics[width=0.9\columnwidth]{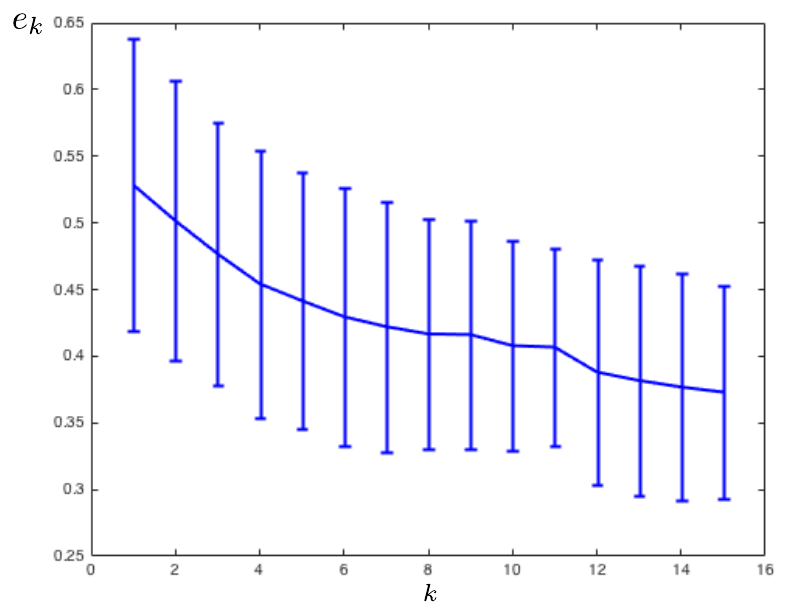}
	\caption{Evolution of the mean and standard deviation of $e_k$ at each iteration $k$ for all $9$ setups and experiments.}
	\label{fig:barErrorAll}
\end{figure}

Since part of the action selection process is randomized, and since the initial and desired images vary from one setup the other, it is difficult to objectively assess the performance of our framework over multiple experiments. To this end, we have processed the results obtained in all $9$ setups (shown in Figures ~\ref{fig:CESexperiments},~\ref{fig:tapExperiments} and~\ref{fig:pushtapExperiments}). First, since the three setups in Fig.~\ref{fig:CESexperiments} have each been tackled with all three pushing strategies (see Fig.~\ref{fig:mutInfoPush}), for each setup we have averaged the three values of $e_k$ (one per strategy) at each iteration $k$. Now, we have $9$ trends of $e_k$ corresponding to the $9$ setups: $3$ for pushing obtained by averaging as mentioned just above, $3$ for tapping and $3$ for pushing+tapping obtained via the experiments of Sections~\ref{Sect.tapExp} and~\ref{Sect.pushtapExp}, respectively. Then, we compute the mean and standard deviation of $e_k$ at each iteration $k$ of these $9$ trends. We do so on the first $15$ iterations, since only $5$ of the $9$ experiments have lasted longer, due to the termination conditions. The results are plotted as error bars in Fig.~\ref{fig:barErrorAll}. As the reader can see, the trend is decreasing, showing that our framework is capable of reducing the mutual information error despite the variability of actions and setups.

\section{Conclusions and future work}\label{sect:conclusion}

In this paper, we have addressed the problem of non-prehensile shaping of plastic materials. 
Inspired by our human study, we have designed two actions, pushing and tapping. Both are realized using image-based visual servoing (neither force nor tactile feedback) to control a tool held by a robot manipulator. We assume that these two actions are specialized: pushing alters the external shape contours, whereas tapping modifies the image. 

The key issue is how to relate the parameters of these two actions to the current and next states. While for tapping this seems simple, since the effect is local and has constant size (equal to the tool contact surface dimensions), pushing requires a deeper reflection, which constitutes in our opinion one of the main contributions of the paper. We draw inspiration from the user dataset to derive the main parameters -- contour and action size -- that humans use when operating. Based on these parameters, we break the global problem of regulating the current image towards the desired one into smaller local problems -- in the state space -- which can be solved more easily.

We also propose three strategies: two heuristics (maximum and average) and a neural network trained with the dataset images. To compare the approaches, we use mutual information -- an objective metric of image similarity. The results show that while slightly outperforming the other strategies on the short run, the neural network quickly settles when the image error is small. Paradoxically, the much simpler maximum strategy is more efficient afterwards (when the error is small). This result is particularly interesting in the light of the current euphoria that surrounds machine learning worldwide. It turns out that for our problem, while requiring a huge pre-processing effort (many images are unusable and deriving push-action triplets is non-trivial, see Sect.~\ref{sect:gettingTriplets}) the outcome of learning is barely better than that of a simple heuristic, because of the curse of dimensionality. This result per se may discourage the user from going through the data acquisition and pre-processing steps. Nevertheless, these steps are also necessary to infer the human parameters mentioned in the paragraph above, which make the global task separable into smaller local ones. Once data acquisition and pre-processing are done, training the neural network is straightforward, so why not use it afterwards? 

All in all, the results show that our framework succeeds in making the robot realize numerous shapes. Nevertheless, we acknowledge the many limitations of our exploratory work. These could be the object of future work of researchers interested by this fascinating topic. A non-exhaustive list is given below to conclude the paper.

\begin{itemize}

\item
The pushing actions clearly outperform the tapping ones. Possible reasons have been mentioned in Sect.~\ref{Sect.tapExp}. The main one seems to be the choice of the feature: an image, rather than a point cloud. Yet, with an accurate depth image, tapping could be formulated as pushing in any plane perpendicular to the image. Then, our approach for pushing could be directly applied in any depth plane and would likely succeed in generalization.

\item 
Many limitations are due to the hardware constraints: tool size, design, sensed data. The use of a soft tool or of multiple tools (e.g., fingers) could improve the performance, while raising other interesting research problems.

\item
In our approach, inspired by model-based control, we map current and desired state to action. Alternatively, one could map current state and action to next state. This second paradigm is more relevant if planning or reinforcement learning were to be applied to this problem.

\item
It would be useful to integrate haptic feedback (measured by force or tactile sensors) to vision. This would be valuable, e.g., to control the force required to overcome the kinetic sand resistance.

\item
We have trained the neural network on the user dataset. One could speed up the learning process via self-learning (\textit{i.e.}, the robot acquires new data while it molds the material).

\item 
A larger set of actions could be studied. For instance, the analysis of the user dataset shows that often humans perform hybrid actions (e.g., simultaneous ``push-and-tap''), which alter multiple features at once. 

\item 
Our user analysis is quite limited (with only nine partakers). While being a world premiere, our public dataset should be enriched by other researchers, particularly cognitive scientists, who could provide their expertise on human studies.

	

\item Our framework is not autonomous in selecting the best action type, given the system state. This problem is of interest not only for deformable object manipulation, but also for other applications that need heterogeneous action sequencing.

\item The current version of our framework does not regulate the tool orientation. This feature could be added in future work, by drawing inspiration from the suggestions given in Sect.~\ref{sec:toolOrient}.

\end{itemize}


\begin{acks}
This work has been partly funded by the PHC FASIC project FlexBot. Ortenzi, Cosgun and Corke are supported by the Australian Research Council Centre of Excellence for Robotic Vision (project number CE140100016). We would like to thank Steven Martin for the help with the custom 3D printed tool, and El Mustapha Mouaddib, Andr\'e Crosnier and Juxi Leitner for the constructive discussions.
\end{acks}

%
%
%

\bibliographystyle{SageH}
\bibliography{library}

\end{document}